\documentclass[journal]{IEEEtran}
\ifCLASSINFOpdf
\else
\fi
\usepackage{subfigure}
\usepackage{float}
\usepackage{multirow}
\usepackage{multicol} 
\usepackage{xcolor}
\usepackage{colortbl}  %
\usepackage{threeparttable} 
\usepackage{enumerate}
\usepackage{amsmath,amssymb}
\usepackage{setspace}  
\usepackage{verbatim}
\usepackage{bm}
\usepackage{algorithm}
\usepackage{algorithmic}
\usepackage{stfloats}
\usepackage{hyperref}
\usepackage{stfloats}
\usepackage{amsmath}
\usepackage{bm}
\usepackage{graphicx} 
\usepackage{textcomp}
\usepackage{subfigure}
\usepackage{color}
\usepackage{array}
\begin{document}
%
\title{3D Face Anti-Spoofing with Factorized Bilinear Coding}
\author{Shan Jia,
        Xin Li,~\IEEEmembership{Fellow,~IEEE,}
        Chuanbo Hu,
        Guodong Guo,~\IEEEmembership{Senior member,~IEEE,}
        and~Zhengquan Xu
\thanks{S. Jia and Z. Xu are with the State Key Laboratory of Information Engineering in Surveying Mapping and Remote Sensing, Wuhan University, Wuhan 430079, China (e-mail: jias@whu.edu.cn; xuzq@whu.edu.cn).}
\thanks{X. Li, G. Guo, and C. Hu are with the Lane Department of Computer Science and Electrical Engineering, West Virginia University, Morgantown, WV 26506, USA (e-mail: \{xin.li, chuanbo.hu, guodong.guo\}@mail.wvu.edu)}
\thanks{Corresponding author: Xin Li.}}

\maketitle

\begin{abstract}
We have witnessed rapid advances in both face presentation attack models and presentation attack detection (PAD) in recent years. When compared with widely studied 2D face presentation attacks, 3D face spoofing attacks are more challenging because face recognition systems are more easily confused by the 3D characteristics of materials similar to real faces. In this work, we tackle the problem of detecting these realistic 3D face presentation attacks and propose a novel anti-spoofing method from the perspective of fine-grained classification. Our method, based on factorized bilinear coding of multiple color channels (namely MC\_FBC), targets at learning subtle fine-grained differences between real and fake images. By extracting discriminative and fusing complementary information from RGB and YCbCr spaces, we have developed a principled solution to 3D face spoofing detection. A large-scale wax figure face database (WFFD) with both images and videos has also been collected as super realistic attacks to facilitate the study of 3D face presentation attack detection. Extensive experimental results show that our proposed method achieves the state-of-the-art performance on both our own WFFD and other face spoofing databases under various intra-database and inter-database testing scenarios.

\end{abstract}

\begin{IEEEkeywords}
Face spoofing attack, presentation attack detection, wax figure face, face anti-spoofing.
\end{IEEEkeywords}

\IEEEpeerreviewmaketitle
\section{Introduction}
\IEEEPARstart{F}{ace} has been one of the most widely used biometric modalities due to its accuracy and convenience for personal verification and identification. However, the increasing popularity and easy accessibility of face modalities make face recognition systems (FRS) a major target of spoofing such as presentation attacks~\cite{jain2004introduction}. This class of security threats can be easily implemented by presenting the FRS a face artifact, which is also known as presentation attack instrument (PAI)~\cite{ISO2016}. A recent breach of biometrics database (BioStar) leads to the compromise of as many as 28 million records containing facial and fingerprint data, which can be exploited by malicious hackers as PAIs.

\begin{figure}[t]
\setlength{\abovecaptionskip}{0pt} 
\setlength{\belowcaptionskip}{0pt} 
\begin{center}
\subfigure[]{\label{(a)}
\includegraphics[width=1.21in,height=0.962in]{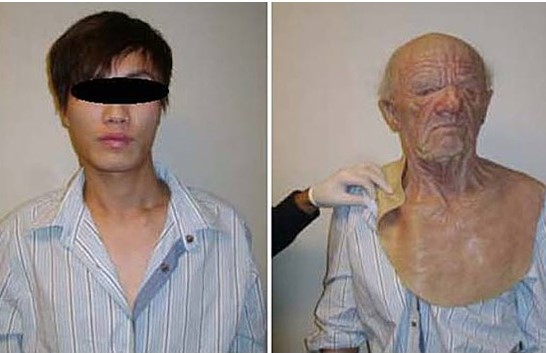}}
\subfigure[]{\label{(b)}
\includegraphics[width=1.0in,height=0.962in]{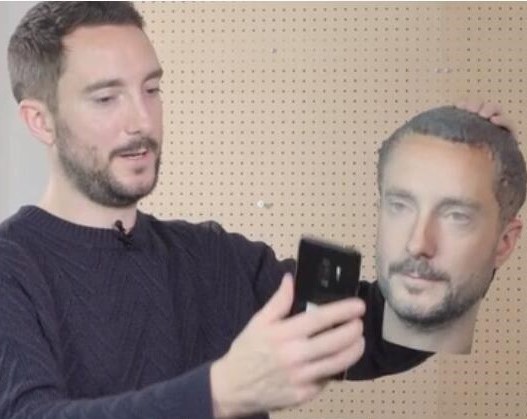}}
\subfigure[]{\label{(c)}
  \includegraphics[width=1.03in,height=0.962in]{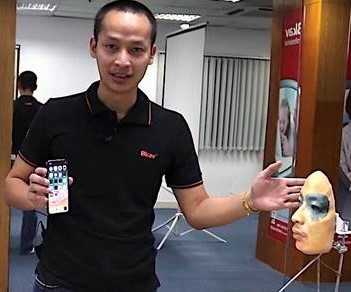}}
\caption{Examples of 3D presentation attack cases. (a) Airport security system fooled by silicon mask\protect\footnotemark[1], (b) Android phones fooled by a 3D-printed head\protect\footnotemark[2], (c) iPhone X face ID unlocked by a 3D mask\protect\footnotemark[3].}
\label{fig:1}  
\vspace{-0.7cm} 
\end{center}
\end{figure}

Based on the way of generating face artifacts, face presentation attacks can be classified into 2D (e.g., printed/digital photographs or recorded videos on mobile devices such as a tablet) and 3D (e.g., by wearing a mask or presenting a synthetic model). Existing research on FRS has paid more attention to 2D face PAI due to its simplicity, efficiency, and low cost. However, as material science and 3D printing technology advance, creating face-like 3D structures or materials has become easier and more affordable. When compared to 2D attacks, 3D face presentation attacks are more realistic and therefore more difficult to be detected. The class of 3D face presentation attacks includes wearing wearable facial masks~\cite{3DMAD2013spoofing}, building 3D facial models~\cite{xu2016virtual}, through facial make-up~\cite{guo2013face, chen2017spoofing}, and using plastic surgery. Fig. \ref{fig:1} shows several examples of 3D presentation attacks, which have successfully fooled some widely used FRS in practice. 

\footnotetext[1]{Picture is downloaded from https://chameleonassociates.com/security-breach/.}
\footnotetext[2]{Picture is downloaded from http://www.floridaforensicscience.com/broke-bunch-android-phones-3d-printed-head/.}
\footnotetext[3]{Picture is downloaded from https://boingboing.net/2010/11/05/young-asian-refugee.html.}

The vulnerability of current face recognition systems to realistic face presentation attacks has facilitated a series of studies on 3D face presentation attack detection (PAD)~\cite{jia2020survey}. Existing methods tried to explore the difference between real face skin and 3D fake face materials based on the reflectance properties using multispectral imaging~\cite{kose2013reflectance, liu2018detecting}, texture analysis~\cite{agarwal2016face, wang2018face}, deep features~\cite{shao2018joint, qin2019learning}, or liveness cues~\cite{liu20163d2, lin2019face}. They have achieved promising detection performance on several existing 3D face spoofing databases~\cite{3DMAD2013spoofing,liu20163d,manja2017detecting,george2019biometric}. 
However, since facial masks are often made of paper, latex or silicone materials, these 3D face spoofing databases have the limitations of small database size (mostly less than 30 subjects), poor authenticity (some based on 2D planar or noncustomized masks~\cite{manja2017detecting, agarwal2017face}), and low diversity in subject and recording process, which might impede the development of effective and practical PAD schemes. Several studies~\cite{jia2020survey, liu2020temporal, shao2017deep, manjani2017detecting} have shown that a variety of 3D PAD methods suffer from performance degradation on databases with more diverse and realistic 3D face spoofing attacks.

\begin{figure}[t]
\setlength{\abovecaptionskip}{0pt} 
\setlength{\belowcaptionskip}{0pt} 
\begin{center}
\subfigure[]{\label{(a)}
\includegraphics[width=1.63in]{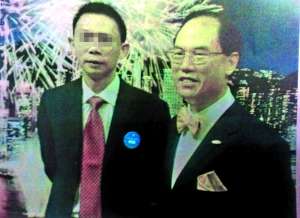}}
\subfigure[]{\label{(b)}
 \includegraphics[width=1.63in]{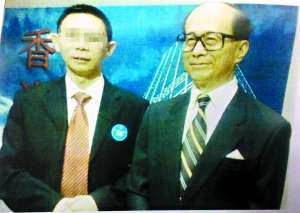}}
\caption{Photos with wax figure faces for fraud. In 2012, six suspects snapped about 600,000 people out of nearly US\$475 million under a pyramid sales scam using photos taken with the wax figures at Hong Kong’s Madam Tussauds Museum, (a) with the wax figure of chief executive Donald Tsang Yam-Kuen, (b) with the wax figure of business tycoon Li Ka-Shing.}
\vspace{-0.64cm} 
\label{fig:2}  
\end{center}
\end{figure}


In this work, we aim at detecting realistic 3D face spoofing attacks which often pose greater threats to existing FRS (even with face anti-spoofing modules) than their 2D counterparts. Based on the fact that wax figure faces have already been used for identity, personation, and fraud in the real world (as shown in Fig. \ref{fig:2}), we first collect a large number of wax figure faces based on both still images and moving videos as super realistic 3D face spoofing attacks by expanding upon our preliminary work~\cite{jia2019database}. 
Furthermore, we propose to treat realistic face spoofing detection as a special class of fine-grained image classification problems which focus on differentiating between hard-to-distinguish object classes. Then we introduce a new bilinear coding based method with state-of-the-art performance of combining features for fine-grained classification, which generates discriminative representations for 3D face anti-spoofing. The main contributions of this work are summarized below.

\begin{itemize}
\item A large-scale wax figure face database (WFFD) with both images and videos is constructed as the super realistic 3D face presentation attack. It contains 2300 pairs of matched faces from 745 subjects (totally 4600 faces), and 285 videos from 241 subjects (with over 45000 frames).

\item A new 3D face anti-spoofing method, named MC\_FBC, is proposed based on factorized bilinear coding for multicolor channels. This is the first work addressing the problem of detecting a realistic face PAD in a fine-grained manner and use bilinear coding to improve the discriminative power of feature representations.
 
\item We have conducted extensive experiments on the proposed WFFD and several publicly available databases. Our findings show that the proposed method outperforms several state-of-the-art methods under both intra- and inter-database testing scenarios.
\end{itemize}

The rest of this paper is organized as follows. In Section II, we briefly review related research in 3D face PAD methods and spoofing databases. Section III introduces the WFFD database with both photo-based and video-based wax figure faces as super realistic 3D face spoofing attacks. The proposed 3D face anti-spoofing method based on factorized bilinear coding is presented in Section IV. Experimental results are reported in Section V, and we make several conclusions about this work and future research in Section VI.
\section{Related work}

\subsection{3D Face PAD methods}

Detection of 3D fake faces is often more challenging than detecting fake faces with 2D planar surfaces. Methods designed for 2D face spoofing detection may fail to identify 3D face spoofing attacks, especially for those using recapture effects~\cite{patel2016secure} and spoofing medium contour detection~\cite{zhu2019detection}. Existing PAD methods for 3D face presentation attacks, mainly based on the difference between real face skin and mask material, can be broadly classified into five categories, namely, reflectance-based, texture-based, shape-based, liveness-based, and deep feature based.

Earlier studies~\cite{kim2009masked, wang2013new, steiner2016reliable} in 3D mask spoofing detection were based on the reflectance difference of facial skin and mask materials. For example, the distribution of albedo values for illumination of various wavelengths was first analyzed in~\cite{kim2009masked} to find how different facial skins and mask materials (silicon, latex, and skinjell) behave in terms of reflectance. 
Texture-based methods explore the texture pattern difference of real faces and masks with the help of texture feature descriptors, such as the widely used Local Binary Patterns (LBP)~\cite{3DMAD2013spoofing,erdog2014spoofing}, Binarized Statistical Image Features (BSIF)~\cite{naveen2016face}, and Haralick features~\cite{agarwal2016face}. 

Shape-based 3D mask PAD methods use shape descriptors~\cite{MORPHO3kose2013vulnerability, tang20173d, hamdan2017detection} or 3D reconstruction~\cite{wang2018face} to extract discriminative features from faces and 3D masks. Different from reflectance-based or texture-based detection methods, these schemes only require standard color images without the need of special sensors. However, their detection performances rely on the quality of 3D mask attacks, and may not be robust to super realistic 3D face presentation attacks.
More recently, some methods explore liveness cues to detect 3D face presentation attacks, such as thermal signatures~\cite{bhat2017you}, gaze information~\cite{alsufyani2018biometrie, ali2018gaze}, and pulse or heartbeat signals~\cite{liu2020temporal,liu2018remote, hern2018time, Li2017Generalized}. Based on the intrinsic liveness signals, these methods achieve outstanding performance on distinguishing real faces from masks.

Instead of extracting hand-crafted features, deep feature based methods automatically extract features from face images. Two deep representation approaches were investigated in~\cite{menotti2015deep} for spoofing detection in different biometric modalities. Image quality cues (Shearlet) and motion cues (dense optical flow) were fused in~\cite{feng2016integration} using a hierarchical neural network for mask spoofing detection, which achieved a Half Total Error Rate (HTER) of 0\% on the 3DMAD database~\cite{3DMAD2013spoofing}. A network based on transfer learning using a pre-trained VGG-16 model architecture is presented in~\cite{lucena2017transfer} to recognize photo, video, and 3D mask attacks. Based on the observation of the importance of dynamic facial texture information, a deep convolutional neural network-based approach was developed in~\cite{shao2018joint, shao2017deep}. Both intra-dataset and cross-dataset evaluation on 3DMAD and their supplementary dataset indicated the efficiency and robustness of the proposed method. 

Despite these advances in 3D face anti-spoofing, there are limitations in each category of detection methods. For example, the main limitation of reflectance-based methods is the requirement of special and expensive devices to acquire multispectral images at varying wavelengths. Although texture and shape-based methods are easy-to-implement, their robustness to different mask spoofing attacks needs further investigation. The liveness cues, such as pulse/heartbeat-based, are highly dependent on lighting conditions and camera settings (e.g., exposure time and frame rates). Deep feature based approaches are generally sensitive to dataset sizes and lack transparency. In addition, most of them~\cite{jia2020survey, liu2020temporal,hernandez2018time, shao2017deep, manjani2017detecting, ali2017biometric, alsufyani2018biometrie, liu2018remote, hern2018time} still suffer from performance degradation when applied to databases with more realistic face spoofing attacks. How to improve the {\em robustness} and {\em discriminative power} of representations to distinguish real faces from fake ones has remained a long-standing open problem.

\vspace{-0.1in}
\subsection{3D Face Spoofing Databases}
Existing 3D face spoofing databases create attacks mainly based on wearable 3D face masks that are made of material with face characteristics similar to real faces. 3DMAD~\cite{3DMAD2013spoofing} is the first publicly available 3D mask database. It used the services of ThatsMyFace\footnote[4]{http://thatsmyface.com/} to manufacture 17 masks of users, and recorded 255 video sequences with an RGB-D camera of Microsoft Kinect device for both real access and presentation attacks. 
With the development of 3D modeling and printing technologies, more mask databases have been created since 2016. 3DFS-DB~\cite{galbally2016three} is a self-manufactured and gender-balanced 3D face spoofing database, in which 26 printed models were made using two 3D printers. 
HKBU-MARs~\cite{liu20163d, liu2020temporal} is another 3D mask spoofing database with more variations. It generated customized masks from two companies (ThatsMyFace and REAL-F\footnote[5]{http://real-f.jp/en\_the-realface.html}), and created videos from 12 subjects. To include more subjects, SMAD database~\cite{manja2017detecting} has collected videos of people wearing silicone masks from online resources. It contains 65 genuine videos of people auditioning, interviewing, or hosting shows, and 65 imposter videos of people wearing a complete 3D (but not customized) mask which fits well with proper holes for the eyes and mouth. 

For effective detection, more 3D face spoofing databases provide multi-modality spoofing samples under special lighting conditions. The BRSU Skin/Face/Spoof Database~\cite{steiner2016reliable} provides multispectral SWIR (Short Wave Infrared) and RGB color images for various types of masks and facial disguises. 
The MLFP database~\cite{agarwal2017face} is another multispectral database using latex and paper masks. It contains 1350 videos of 10 subjects in visible, near infrared (NIR), and thermal spectrums. The ERPA database~\cite{bhat2017you} provides RGB and NIR images of both bona fide and 3D mask attack presentations captured using special cameras. This is a small dataset with only 5 subjects involved. Both rigid resin-coated masks and flexible silicone masks are considered. Similarly, the recently released WMCA database~\cite{george2019biometric} used multiple capturing devices/channels, including color, depth, thermal, and infrared. It contains 1679 videos with 347 bona fide and 1332 attacks from 72 subjects; among them, 709 3D face spoofing attack videos include fake head, rigid mask, flexible silicone masks, and paper masks. The 3DMA~\cite{xiao20193dma} also collects 920 videos captured in both visual and NIR modalities from 67 subjects.

\begin{figure}[t]
\setlength{\abovecaptionskip}{0pt} 
\setlength{\belowcaptionskip}{0pt} 
\begin{center}
\includegraphics[width=3.45in]{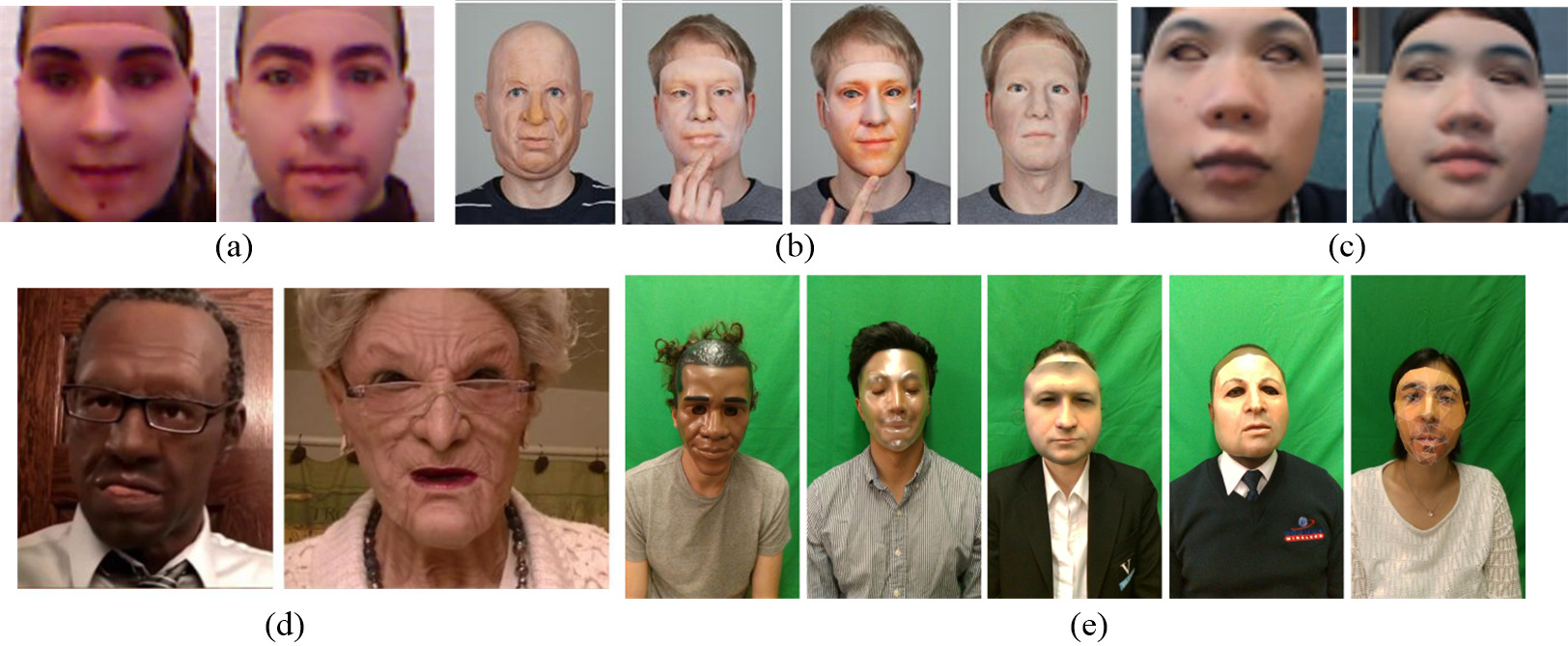}
\caption{3D mask spoofing examples in existing databases. (a) 3DMAD, (b) 3DFS-DB, (c) HKBU-MARs, (d) SMAD, (e) WMCA.}
\label{fig:2a}  
\vspace{-0.6cm} 
\end{center}
\end{figure}

These databases have played a significant role in designing multiple detection schemes against 3D face presentation attacks. However, the following issues have remained to be addressed: (1) only a small number of 3D spoofing samples are collected due to the difficulty and expense of mask production; (2) most of the 3D face spoofing attacks are of low quality (e.g., using noncustomized masks~\cite{manja2017detecting, agarwal2017face} or 2D cut-out paper masks~\cite{agarwal2017face, li2018unsupervised}. Although soft/flexible masks are closer to real faces, they are still easy to be recognized by humans~\cite{3DMAD2013spoofing, liu20163d, george2019biometric, galbally2016three} (as shown in Fig. \ref{fig:2a}); (3) they contain less variations to simulate the real world scenarios in terms of recording cameras, lighting settings, subject pose/age, and facial expression/resolution. Therefore, we have introduced wax figure faces as super realistic 3D face spoofing attacks in~\cite{jia2019database} (with 4400 still faces). After that, Vareto et al.~\cite{vareto2019swax} collected a wax figure database named SWAX with 1800 faces and 110 videos from 55 people/waxworks. 

In this paper, to overcome the limitations in both 3D face spoofing databases and detection methods, we first expand the wax figure face dataset to include both images and videos with a high degree of diversity, aiming at more faithfully simulating realistic spoofing attacks. Then we will design a novel detection method based on the fusion of complementary skin-inspired features in different color spaces to extract subtle differences between real faces and realistic spoofing attacks.
\section{Wax Figure Face Database}
In this section, we introduce the super realistic Wax Figure Face Database (WFFD) with both photo-based and video-based wax figure faces to address the weaknesses in the existing 3D face spoofing databases. Considering the difficulty and expense in generating a large number of 3D wax figure faces, we take advantage of the popularity and publicity of numerous celebrity wax figure museums in the world, and collect as many wax figure faces as possible from online resources to construct such a database with a large size and high diversity. These wax figure faces are all carefully designed and made in clay with wax layers, silicone, or resin materials, so that they are super realistic and similar to real faces. In the physical domain, the attacker may fool the FRS using just the life-size wax heads, which can be obtained with relatively low cost (about 200 dollars) from online shopping websites (such as eBay US and Taobao China).

\subsection{Wax figure face database of still images} 
\begin{figure}[t]
\setlength{\abovecaptionskip}{-4pt} 
\setlength{\belowcaptionskip}{0pt} 
\begin{center}
\includegraphics[height=3.9cm]{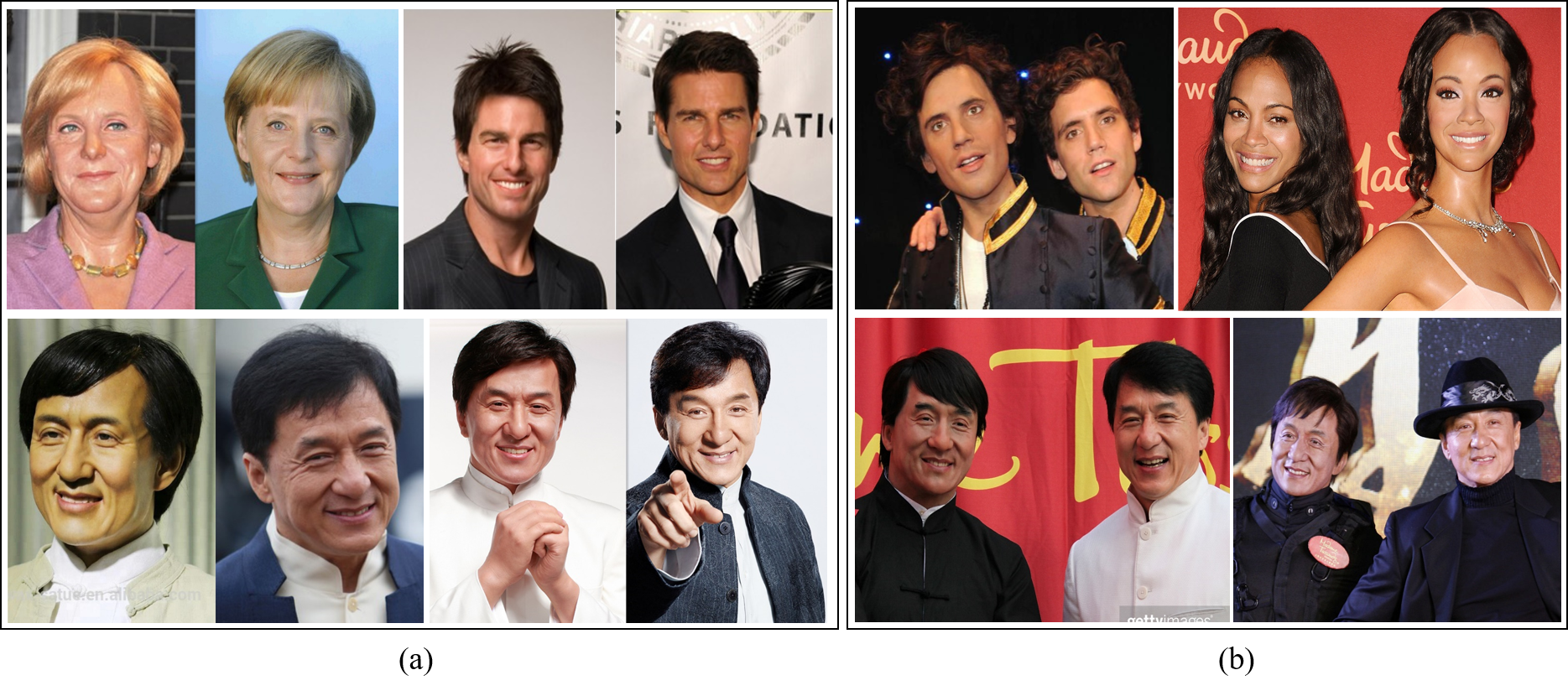}
\caption{Examples in the WFFD database. (a) in Protocol I, (b) in Protocol II. Note that one subject may have several wax figures in each protocol.}
\label{fig:3a}    
\vspace{-0.6cm}
\end{center}
\end{figure}

Our previous work~\cite{jia2019database} has been greatly expanded to qualify wax figure faces as super realistic 3D face spoofing attacks. We have thoroughly cleaned the dataset and removed images with tiny faces or low quality due to lossy compression, and then added 100 new pairs of images from 75 subjects to the new database. Finally, a total of 2300 pairs of images (4600 faces with both real and wax figure faces) of 745 subjects are collected. We have followed the three protocols designed in~\cite{jia2019database}: Protocol I with wax figure faces and real faces grouped manually from different devices and environmental conditions (e.g., Fig. \ref{fig:3a}(a)); Protocol II with wax figure and real faces recorded in the same environment with the same camera (e.g., Fig. \ref{fig:3a}(b)); and Protocol III combining the previous two protocols to simulate real-world operational scenarios. Table \ref{tab:1} provides the details about the statistics of images, faces, and subjects in each protocol. 

\vspace{-0.2cm}
\begin{table}[h]
\setlength{\abovecaptionskip}{0.cm}
\setlength{\belowcaptionskip}{-1.1cm}
\newcommand{\tabincell}[2]{\begin{tabular}{@{}#1@{}}#2\end{tabular}}
\small
\renewcommand{\arraystretch}{1.2}
\setlength{\tabcolsep}{4pt}
\caption{Details of each protocol in the WFFD}
  \centering
    \begin{tabular}{lcccccc}
\hline        
    \multirow{2}[3]{*}{\textbf{Protocol}} & \multicolumn{4}{c}{\textbf{\#Image}} & \multirow{2}[3]{*}{\textbf{\#Face}}&  \multirow{2}[3]{*}{\textbf{\#Subject}}\\
\cline{2-5}          & \textbf{Train} & \textbf{Valid} & \textbf{Test} & \textbf{Total} \\
     \hline
    Protocol I & 600   & 200   & 200   & 1000  & 2000  & 462 \\
     \hline
    Protocol II & 780   & 260   & 260   & 1300  & 2600  & 409 \\
     \hline
    Protocol III & 1380  & 460   & 460   & 2300  & 4600  & 745 \\
     \hline
    \end{tabular}%
  \label{tab:1}%
\begin{tablenotes}
\scriptsize
\item[] Note that the train, validation, and test subsets are non-overlapped.
\end{tablenotes}
\end{table}

\vspace{-0.4cm}
\begin{figure}[h]
\setlength{\abovecaptionskip}{0pt} 
\setlength{\belowcaptionskip}{0pt} 
\begin{center}
\includegraphics[width=3.46in]{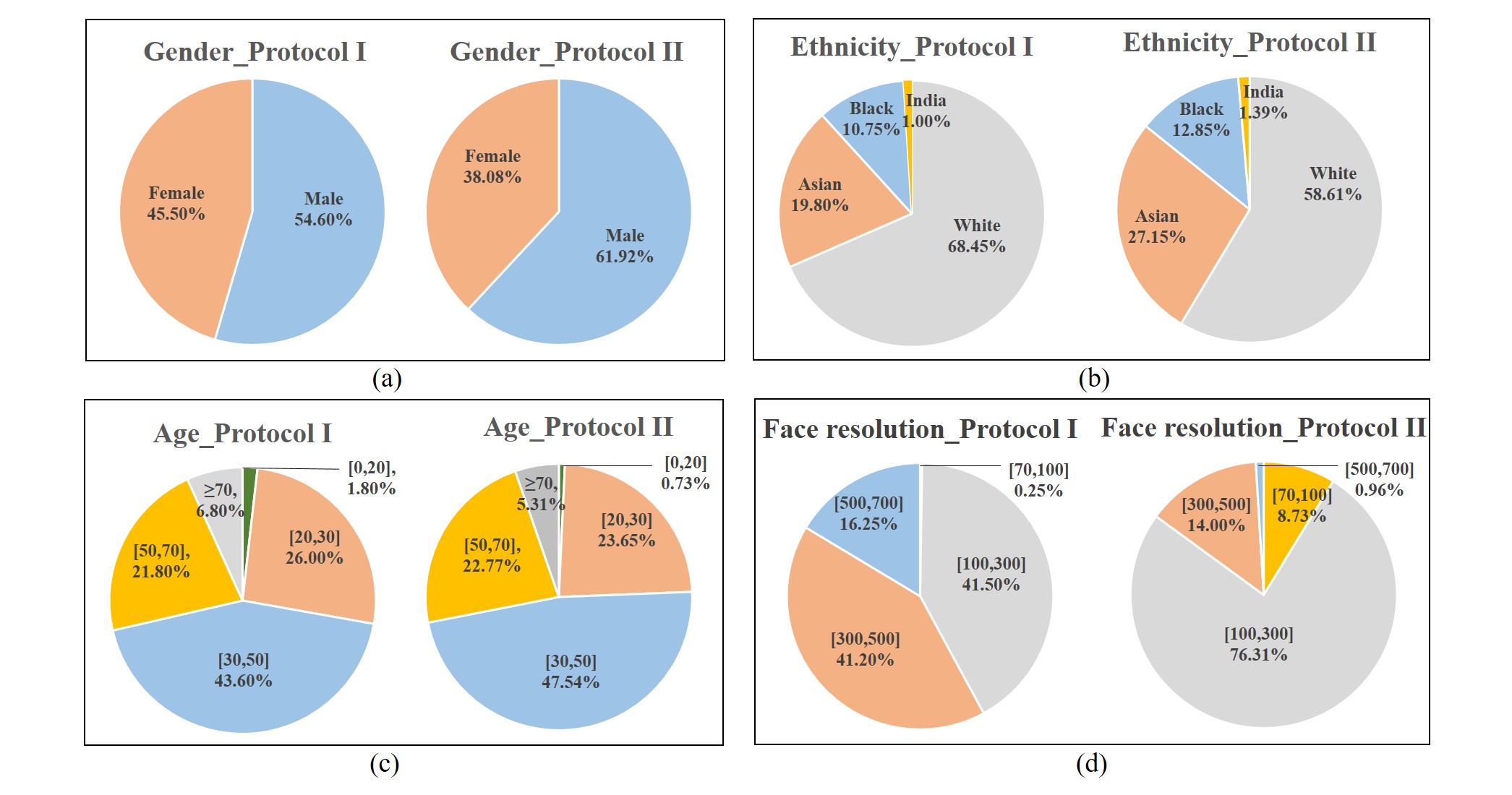}
\caption{Statistical distribution of the WFFD. (a) Gender, (b) ethnicity, (c) age, (d) face resolution.}
\label{fig:4}  
\vspace{-0.3cm} 
\end{center}
\end{figure}

The statistical information about subject gender, age, ethnicity (detected by Face++\footnote[6]{https://www.faceplusplus.com/face-compare-sdk/}), and face resolution (cropped by the dlib face detector~\cite{king2009dlib}) in the WFFD is shown in Fig. \ref{fig:4}. It can be seen that the images in the WFFD are relatively gender balanced - with about 60\% of males and 40\% of females in both protocols. The ethnicity distribution in Fig. \ref{fig:4}(b) contains a majority of White subjects (around 60\%), followed by about 20\% Asians and 10\% Blacks, and a small percentage of Indians (no more than 2\%). We can also see a wide distribution of ages in Fig. \ref{fig:4}(c). The two protocols have similar distribution patterns in terms of age, with half subjects being between 30 and 50 years old. Although the dimensions of most face regions are between $100\times100$ and $500\times500$, there is a big difference in the distribution between the first two protocols. Matched and grouped manually, the dimensions of face regions in Protocol I are generally larger than those in Protocol II. Additionally, the images in Protocol I are more diversified in terms of subject poses, facial expression, recording environment, and devices than those in Protocol II.

\vspace{-0.1in}

\subsection{Wax figure face database of videos}
Inspired by the new generation of intelligent or robotic wax figures which can move and interact with visitors (like the one unveiled by Madame Tussauds in Shanghai in 2018), we have further collected video-based wax figure faces from online resources to obtain moving wax figure faces. 
Similar to the collection of still wax figure faces, we first download as many short videos with celebrity wax figure faces as possible. Then we clean the dataset manually based on the selection criterion that videos without frontal faces, or with faces but containing a significant amount of occlusion or embedded text, should be excluded from the dataset \cite{cevikalp2019face}. Finally, a total of 145 wax figure videos, as well as 140 real face videos have been collected as the newly-constructed WFFD-V database. All real and fake videos are between 60 and 420 frames in length. More details of the WFFD-V with 285 gender-balanced videos are included in Table \ref{tab:videos}. They are randomly partitioned into non-overlapped training, validation, and testing subsets in the same ratio of 3:1:1 as the WFFD still image dataset for performance evaluation. 

\begin{table}[h]
\setlength{\abovecaptionskip}{0.cm}
\setlength{\belowcaptionskip}{-1.1cm}
\renewcommand{\arraystretch}{1.2}
\newcommand{\tabincell}[2]{\begin{tabular}{@{}#1@{}}#2\end{tabular}} 
  \centering
  \small
  \caption{Detailed characteristics of newly constructed WFFD-V dataset.}
    \begin{tabular}{cccccccc}
    \hline
\multirow{2}{*}{\textbf{Video}} & \multicolumn{3}{c}{\textbf{\#Subject}} & \multicolumn{3}{c}{\textbf{\#Video}} & \multirow{2}{*}{\textbf{\#Frame}} \\
\cline{2-7}          & \textbf{F} & \textbf{M} & \textbf{Total} & \textbf{F} & \textbf{M} & \textbf{Total} &  \\
    \hline
    Real  &  53   &  74  & \textbf{127} &58  & 82    & \textbf{140} & 28970 \\
    \hline
    Wax  &  47  &  76   & \textbf{123} & 61    &   84  & \textbf{145} & 16765 \\
    \hline
    Total &  96 &  145  & \textbf{241} &  119   &   166  & \textbf{285} & 45735 \\
    \hline
    \end{tabular}%
  \label{tab:videos}%
  \begin{tablenotes}
\scriptsize
\item[] F denotes Female, and M denotes Male.
\end{tablenotes}
\vspace{-0.3cm}
\end{table}%

\section{3D face anti-spoofing based on factorized bilinear coding}

With materials and shapes highly similar to real faces, 3D face spoofing attacks often lead to performance degradation of existing face PAD methods \cite{erdog2014spoofing}. To detect realistic 3D spoofing attacks, we propose to explore their subtle differences from real faces based on generating discriminative features in a fine-grained manner. Inspired by recent advances in fine-grained classification \cite{kong2017low,he2019fine,zhang2019fine,zhang2018blind}, and subtle variation detection \cite{shi2019atrial}, we propose to tackle the problem of realistic face spoofing detection by combining the skin color model \cite{phung2002novel} with factorized bilinear pooling \cite{FBC2020}. 
Originally proposed for fine-grained visual recognition, bilinear pooling~\cite{lin2015bilinear} has become a popular tool for multimodal data fusion due to its superiority in exploiting higher-order information among complementary features. In this section, we first construct complementary skin-inspired features in color space and then present a novel method of factorized bilinear coding combining features extracted in different color spaces. 
\vspace{-0.1in}
\subsection{Skin color model}
Color spaces play an important role in image processing and computer vision applications. RGB is the most widely used color space for sensing, representing, and displaying color images. However, due to the high correlation among the three color components (red, green, and blue), RGB color space representation is not necessarily the most appropriate choice for face anti-spoofing. Alternative color-space representations such as luminance/chrominance and hue/saturation are also valid and competing choices. Instead of fusing features from a single RGB color space as most previous bilinear pooling schemes~\cite{FBC2020, lin2015bilinear, gao2016compact, yu2017multi, zhang2018blind} did, we propose to take multiple color spaces into account and extract more discriminative features by combining multicolor space representations for 3D face anti-spoofing. 

Different from RGB color space, YCbCr color space encodes a color image similar to human eyes' retina, which separates the RGB components into a luminance component (Y) and two chrominance components (Cb as blue projection and Cr as red projection; for an analog version, as U and V, respectively). YCbCr space is effective for color feature extraction and has also achieved promising performance in face-related applications (e.g., human face detection \cite{phung2002novel}, 2D face spoofing detection~\cite{A12boul2015face}, and skin classification~\cite{brancati2016dynamic}). 

Considering the high similarity between 3D face spoofing attacks and real faces, we propose to take both RGB and YCbCr color spaces of the face image as the input of CNN-based feature extraction module to obtain robust facial color texture descriptions. {{The key motivation behind combining different color textures is two-fold. First, each {\em color-texture} based analysis can help capture the artifacts of spoofing attacks for detection. The artificial face in face spoofing attacks is either made of special materials different from human skins (e.g., silicon, latex, and skinjell mask) or passes through different camera systems or printing systems (or display devices in 2D attacks)~\cite{boulkenafet2018generalization}. Therefore, artificial face images are likely to suffer from different kinds of quality degradation issues - e.g., the face production material artifacts, the PAI dependent color variations, and limited color reproduction in 2D print or replay spoofing, which do not occur in real faces. Second, the fusion of two different color spaces, namely, the RGB with highly correlated color components, and YCbCr with separated luminance and chrominance components, has the potential of learning {\em complementary} and robust subtle features for face anti-spoofing (see the results in Section V-B and V-G).}}

\vspace{-0.1in}
\subsection{Factorized bilinear coding}
Bilinear pooling was introduced in~\cite{lin2015bilinear} to provide robust image representation for fine-grained image classification. In bilinear pooling models, two feature vectors are fused by an outer product (or Kroneker product of matrices); this way, all pairwise interactions among the given features are considered as follows:

\begin{equation}
\boldsymbol Z=\sum_{(i,j\in \mathbb{S})}\boldsymbol{x_}i \boldsymbol{y_}j^{\top}
\end{equation}
where $\left\{\boldsymbol{x_}i|\boldsymbol{x_}i \in \mathbb{R}^p, i \in \mathbb{S} \right\}$, $\left\{\boldsymbol{y_}j|\boldsymbol{y_}j \in \mathbb{R}^q, j\in \mathbb{S} \right\}$ are two input features, $\mathbb{S}$ is the set of spatial locations (combinations of rows and columns), and $\boldsymbol Z\in \mathbb{R}^ {p\times q} $ is the fused feature descriptor. It can be seen that the size of the bilinear feature descriptor can be large, which makes it computationally infeasible. To generate more compact representations, we have employed the factorized bilinear coding (FBC)~\cite{FBC2020} to more computationally efficiently integrate the features from both RGB and YCbCr color spaces for 3D face anti-spoofing. 
\begin{figure*}[t]
\setlength{\abovecaptionskip}{0pt} 
\setlength{\belowcaptionskip}{0pt} 
\begin{center}
\includegraphics[width=6.6in]{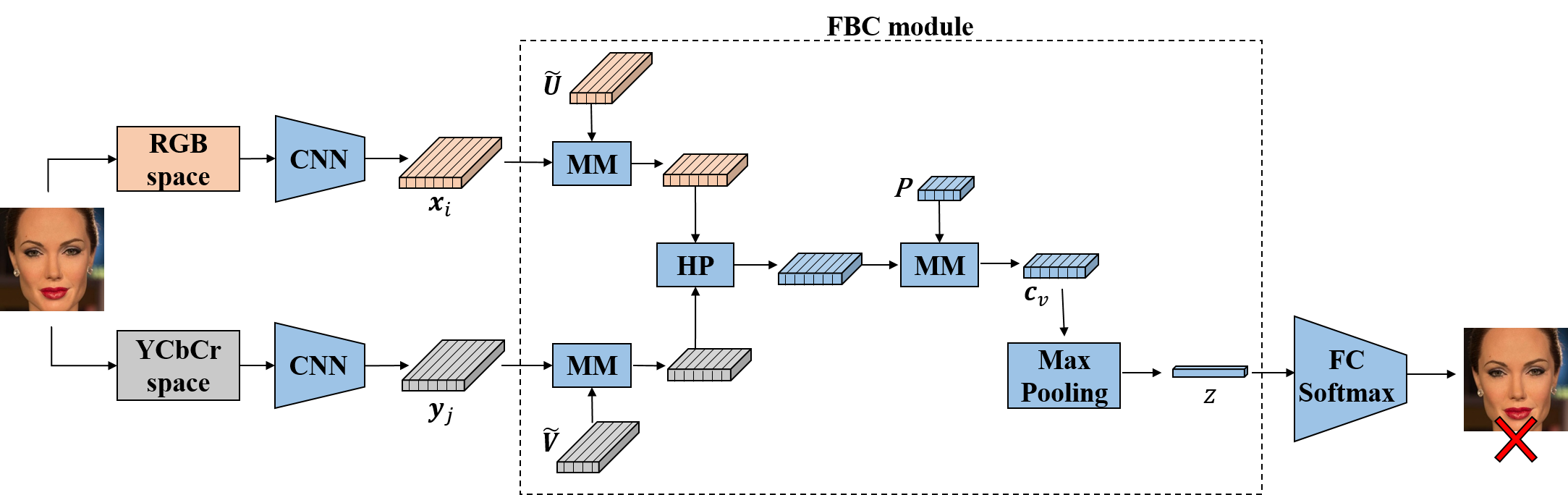}
\caption{Flowchart of the proposed MC\_FBC scheme for 3D face anti-spoofing. {{The FBC module generates discriminative representations in a fine-grained manner}}. MM module refers to 'Matrix Multiplication', and HP module refers to 'Hadamard Product' operation. {{The variables $\boldsymbol{x_}i$ and $\boldsymbol{y_}j$ are the features learned from CNN, $\boldsymbol{\widetilde U}$ and $\boldsymbol{\widetilde V}$ are learnable parameters, $\boldsymbol{P}$ is a fixed binary matrix with only elements in specific rows and columns being ``1", $\boldsymbol{c}_{v}$ is the FBC code, and $\boldsymbol{Z}$ is the final global representation.}}}
\label{fig:3e}
\vspace{-0.6cm}
\end{center}
\end{figure*}

Let $\boldsymbol{x_}i$, $\boldsymbol{y_}j$ be the two features extracted from RGB and YCbCr color spaces respectively, the FBC encodes the features based on sparse coding, and learns a dictionary $B$ with $k$ atoms that are factorized into low-rank matrices to capture the structure of the whole data space. Specifically, let the dictionary $B=[\boldsymbol{b_}1, \boldsymbol{b_}2, ..., \boldsymbol{b_}k] \in \mathbb{R}^{pq\times k} $, and FBC proposes to factorize each dictionary atom $\boldsymbol{b_}l\in \mathbb{R}^{pq}
(1\leq l\leq k)$ into $\boldsymbol{U_}l\boldsymbol{V_}l^{\top}$, where $\boldsymbol{U_}l \in \mathbb{R}^{p \times r}$ and $\boldsymbol{V_}l\in \mathbb{R}^{q \times r}$ are learnable low-rank matrices (as the hyper-parameter rank $r\ll p,q$). Therefore, the original bilinear features $\boldsymbol{x_}i \boldsymbol{y_}j^{\top}$ can be reconstructed by
$\sum\limits_{l=1}^{k}{c_{v}^{l}}{{\boldsymbol{U}}_{l}}\boldsymbol{V}_{l}^{\top}$, with $\boldsymbol{c_}v \in \mathbb{R}^k$ being the FBC code, and ${c_{v}^{l}}$ being the $l$-th element of $\boldsymbol{c_}v$ ($1\leq v\leq N$, $N$ is the number of pairs in $\mathbb{S}$). Then the sparsity-based FBC encodes the two input features $(\boldsymbol{x_}i, \boldsymbol{y_}j)$ into $\boldsymbol{c_}v$ by solving the following optimization problem,
\begin{equation}
\underset{{{\boldsymbol{c}}_{v}}}{\mathop{\min }}\,\bigg|\bigg|{{\boldsymbol{x}}_{i}}\boldsymbol{y}_{j}^{\top}-\sum\limits_{l=1}^{k}{c_{v}^{l}}{{\boldsymbol{U}}_{l}}\boldsymbol{V}_{l}^{\top}\bigg|{{\bigg|}^{2}}+\lambda||{{\boldsymbol{c}}_{v}}|{{|}_{1}}
\end{equation}
where $\lambda $ is a trade-off parameter between the reconstruction error and the sparsity. To obtain the FBC code $\boldsymbol{c}_{v}$, the classical LASSO method~\cite{tibshirani1996regression} has been adopted as shown in Eq.~(3).
\begin{equation}
\left\{ \begin{array}{l}
  \boldsymbol{{c}'}_{v}=\boldsymbol{P}({{\boldsymbol{\widetilde U}}^{\top}}{\boldsymbol{x}_{i}}\circ{{\boldsymbol{\widetilde V}}^{\top}}{{\boldsymbol{y}}_{j}}), \\ 
 {{\boldsymbol{c}}_v}={\rm sign}({\boldsymbol{c}'_v})\circ {\rm max}(({\rm abs}({\boldsymbol{c}'_{v}})-\frac{\lambda }{2}),0). \\
\end{array} 
\right.
\end{equation}
where $\circ$ denotes the Hadamard product, $\boldsymbol{P}\in \mathbb{R}^{k\times rk}$ is a fixed binary matrix with only elements in the row $l$, columns $((l-1)r+1)$ to $(lr)$ being ``1", and $\boldsymbol{\widetilde U}$ and $\boldsymbol{\widetilde V}$ are the transformations of $\boldsymbol{U}$ and $\boldsymbol{V}$ to avoid matrix inversion with heavy computation in the original LASSO method. They are in the form of
\begin{equation}
\left\{ \begin{array}{l}
  {\boldsymbol{\widetilde U}^{\top}}=[\boldsymbol{\widetilde U}_{l}^{\top}] =[\frac{1}{r}\cdot{\boldsymbol{I}}(({\boldsymbol{q}_{l}}\boldsymbol 1_{rk}^{\top})\circ{{\boldsymbol U}^{\top}})]\in {{\mathbb{R}}^{rk\times p}}\\
  {\boldsymbol{\widetilde V}^{\top}}=[\boldsymbol{\widetilde V}_{l}^{\top}] =[\frac{1}{r}\cdot {\boldsymbol{I}}{{\boldsymbol{V}}^{\top}}]\in {\mathbb{R}^{rk\times q}}
\end{array} 
\right.
\end{equation}
where $\boldsymbol{I}\in {\mathbb{R}}^{r\times rk}$ is an all ``1" matrix, $\boldsymbol{q}_{l}$ is the $l$-th column of $\boldsymbol Q=((\boldsymbol P(\boldsymbol U^{\top}{\boldsymbol U}\boldsymbol P^{\top}\cdot \boldsymbol V^{\top}{\boldsymbol V}\boldsymbol P^{\top}))^{-1}{\boldsymbol P})^{\top}$. 

With Eq.~(3), the FBC code $\boldsymbol{c}_{v}$ can be obtained by learning $\boldsymbol{\widetilde U}$ and $\boldsymbol{\widetilde V}$ instead of $\boldsymbol{U}$ and $\boldsymbol{V}$. We can get all FBC codes $\boldsymbol{c} $ in feature pair set $\mathbb{S}$; then they are fused using the max operation to attain the final global representation $z=max\left\{\boldsymbol{c}_{v}\right\}_{i=1}^{N}$. The FBC module is applied to the features extracted by the last convolutional layer of a CNN (e.g., VGG), then followed by a fully connected (FC) layer for classification using Softmax classifier. The whole process of the proposed MC\_FBC scheme is shown in Fig. \ref{fig:3e}. 

\vspace{-0.1in}
\subsection{Loss Function}
To train the network, we utilize the focal loss function~\cite{lin2017focal} which reshapes the standard cross entropy loss in such a way that the loss assigned to well-classified examples in binary classification is downweighted. There are two compelling reasons for such choice of loss function: 1) the higher similarity of real faces and realistic 3D spoofing attacks in 3D face spoofing will lead to harder examples with large errors while the local loss function focuses on training on hard negatives and reducing the loss contribution from easy examples; 2) most 3D face spoofing databases have the problem of class imbalance (e.g., more real samples than fake ones due to the difficulty of collecting 3D spoofing attacks at a large scale). 


Overall, our scheme is different from~\cite{FBC2020} in the following two aspects. First, to get more discriminative color features, we extract features from two different complementary color spaces. Such diversity in terms of feature representation is important to 3D face anti-spoofing, which is sometimes even challenging for human-based inspection. Second, we replace the original cross-entropy loss function with the robust focal loss function to train the network. Originally designed for dense object detection in~\cite{lin2017focal}, we have found that in the scenario of anti-spoofing where training on a sparse set of hard examples is common, it is important to prevent the majority of easy negatives from dominating the detector. Such observation contributes to the improved training performance of new loss functions over the traditional ones. 


\section{Experimental Results}
In this section, we conduct extensive experiments to evaluate our method. We first explore the influence of color spaces on detection performance using the super realistic 3D face spoofing database. Different feature fusion schemes are also compared with our MC\_FBC method. Then we present the comparison results under intra-database testing on the WFFD database and several 2D/3D face spoofing databases, and finally the performance under inter-database testing is evaluated to show the generalizability of the proposed method.

\vspace{-0.1in}
\subsection{Experimental settings}
\subsubsection{Implementation} {{Two backbone networks pretrained on ImageNet~\cite{deng2009imagenet} were fine-tuned on face spoofing datasets to extract features for the FBC module, namely a relatively small VGG-16 model~\cite{simonyan2014very} and ResNet-50~\cite{he2016deep} (deeper and more accurate). It is worth noting that the MC\_FBC can be also applied to other network structures for further improvement of detection performance. The reasons why we used VGG-16 and ResNet-50 models are not only for their robust performance in different detection tasks, but also for a fair comparison with other bilinear pooling based methods where the two models, especially the VGG-16, are widely used~\cite{FBC2020, lin2015bilinear, gao2016compact, zhang2018blind}.}} The last pooling layer and the fully-connected layers were removed in both networks. The learnable parameters $\boldsymbol{\widetilde U}$ and $\boldsymbol{\widetilde V}$ were updated by the back-propagation algorithm to get the FBC code. We set the rank of $\boldsymbol{\widetilde U}$ and $\boldsymbol{\widetilde V}$ as one, the number of dictionary atoms $k$ as 2048, and $\lambda$ as 0.001 (according to~\cite{FBC2020}). As for the training parameters, we fine-tuned the model using SGD with an initial learning rate of 0.01 (decreased by a factor of 10 for every 40 epochs until it reaches 0.0001), weight decay as $5\times {10^{-4}}$, momentum as 0.9, and batch size as 16 for VGG-16 model and 8 for ResNet-50 model. The weighting factor and tune-able focusing parameter in the focal loss function were all set to one. Input images are the cropped faces based on the {\em dlib} face detector~\cite{king2009dlib} with size of $224\times224$. All experiments are conducted using PyTorch on a workstation with Titan XP GPUs.

\subsubsection{Databases} In addition to our WFFD database, we have used two publicly available 3D face spoofing databases - namely, 3DMAD~\cite{3DMAD2013spoofing} (the most widely-used) and HKBU-MARs-V1+ \cite{liu2020temporal} (with hyper-real 3D masks). As both databases contain videos of 300 frames (3DMAD with 255 videos and HKBU-MARs-V1+ with 180), 20 frames were randomly selected for spoofing detection. The averaged scores of these frames were computed as the final score. We have followed the leave-one-out cross-validation (LOOCV) protocol settings for the two databases as previous works did~\cite{shao2018joint, liu2020temporal, liu2018remote}. We have conducted 20 rounds of LOOCV with each round randomly selecting 10 subjects for training and 6 for validation on 3DMAD, while 5 for training and 6 for validation on HKBU-MARs-V1+ database. To validate the generalization property, we also considered two 2D face spoofing databases with both printed photo attacks and replayed video attacks: MSU-USSA~\cite{patel2016secure} and Oulu-NPU database~\cite{boulkenafet2017oulu}. MSU-USSA database was specifically created to simulate face spoofing attacks with diversities of environment, image quality, and acquisition device. It consists of 9,000 images (1,000 live subjects and 8,000 spoofing attacks) recorded with two types of cameras. A five-fold subject-exclusive cross-validation protocol was designed for this database. Oulu-NPU database contains 4,950 videos of 55 subjects with both real access and 2D face spoofing attacks. The videos were recorded using six mobile devices in three sessions with different illumination conditions, and they were divided into three subsets for training, validation, and testing, with four protocols. 

\subsubsection{Evaluation metrics} We report all experimental results following the ISO/IEC 30107-3 metrics~\cite{ISO2017}. Three types of errors, i.e., Attack Presentation Classification Error Rate (APCER), Bona Fide Presentation Classification Error Rate (BPCER), and Average Classification Error Rate (ACER) are used in addition to the detection accuracy. 
\begin{table*}[htbp]
\setlength{\abovecaptionskip}{0.cm}
\setlength{\belowcaptionskip}{-1.1cm}
  \centering
  \small
  \caption{Comparison results (\%) of different color spaces on WFFD database under Protocol III}
    \begin{tabular}{l|cccc|cccc}
    \hline
    \multirow{2}[0]{*}{\textbf{Color space}} & \multicolumn{4}{c|}{\textbf{Backbone-VGG-16}} & \multicolumn{4}{c}{\textbf{Backbone-ResNet-50}} \\
\cline{2-9}          & \textbf{Accuracy} & \textbf{APCER} & \textbf{BPCER} & \textbf{ACER} & \textbf{Accuracy} & \textbf{APCER} & \textbf{BPCER} & \textbf{ACER} \\
    \hline
    RGB   &  93.26 & 8.91 & 4.57 & 6.74 & 93.59&  8.91&  3.91 & 6.41\\
\hline
    YCbCr & 92.39 & 8.26 & 6.96 & 7.61 &93.91 & 8.70& \textbf{3.48} & 6.09\\
\hline
    YUV   & 87.83 & 11.09 & 13.26 & 12.18 & 92.93 & 9.78 & 4.35 & 7.07\\
\hline
    HSV   &  87.07 &  16.30  &  9.57 & 12.94  & 91.74&10.65 & 5.87 & 8.26\\
\hline
    YUV+YCbCr & 91.09 & 10.87 & 6.96 & 8.92 &93.37 &5.87 & 7.39 & 6.63\\
\hline
    RGB+HSV & 93.15 & 8.26 & 5.43 & 6.85 & 93.91 & 6.30 & 5.87 & 6.09\\
\hline
    RGB+YUV & 94.24 & 7.17 & \textbf{4.35} & 5.76 &94.89 &5.87 & {4.35} & 5.11\\
\hline
    RGB+YCbCr (MC\_FBC) & \textbf{94.57} & \textbf{6.09} & {4.78} & \textbf{5.44} & \textbf{95.22} & \textbf{5.65} & {3.91} & \textbf{4.78}\\
\hline
    \end{tabular}%
  \label{tab:color}%
\vspace{-0.3cm}
\end{table*}%
\vspace{-0.1in}

\subsection{Ablation Study}
\subsubsection{The impact of multiple color spaces}
We first demonstrate the effectiveness of combining multiple color spaces in detecting wax figure faces from real ones on the WFFD dataset. Table \ref{tab:color} shows the comparison results under Protocol III with two features both from the same color space (including RGB, YCbCr, YUV, and HSV) and from two different color spaces as the input of FBC module. It can be observed that in single color space, RGB and YCbCr get higher classification accuracy and lower error rates than YUV and HSV color space. However, using multiple color spaces, especially combining the RGB with YCbCr, obtains better results. Specifically, the proposed MC\_FBC scheme combining RGB and YCbCr color spaces achieved 94.57\% accuracy and 5.44\% ACER under the VGG-16 model, while the performance was further improved by the deeper ResNet-50 model, with the highest accuracy of 95.22\% and the lowest ACER of 4.78\%. This shows the complementary properties of RGB and YCbCr spaces in generating more discriminative representations. 

\subsubsection{Exploring different feature fusion schemes}
We next compare the proposed MC\_FBC scheme with several feature fusion schemes, including concatenation (simply combining two feature vectors by concatenating them), {{score-level fusion of two color spaces,}} traditional bilinear pooling (BP)~\cite{lin2015bilinear}, compact bilinear pooling (CBP)~\cite{gao2016compact}, and factorized bilinear coding (FBC)~\cite{FBC2020}. As all these bilinear pooling based methods fused features from VGG-16 model, we present the comparison results in Table \ref{tab:fusion} using VGG-16 network as the backbone. We can see that without bilinear pooling fusion, the learned features from VGG-16 model suffer from a high ACER of over 14\%. By contrast, bilinear pooling based methods with richer information have dramatically improved the detection performance - i.e., improving the classification accuracy by over six percentage points and reducing the three error rates (APCER/BPCER/ACER) by more than half on average. 

\begin{table}[tbp]
\setlength{\abovecaptionskip}{0.cm}
\setlength{\belowcaptionskip}{-1.1cm}
  \centering
  \small
  \caption{Comparison results (\%) of different fusion methods on WFFD database under Protocol III using VGG-16}
    \begin{tabular}{lcccc}
    \hline
    \textbf{Fusion scheme} & \textbf{Accuracy} & \textbf{APCER} & \textbf{BPCER} & \textbf{ACER} \\
\hline
    Original VGG-16 & 84.68 & 13.48 & 17.17 & 15.33 \\
\hline
    Concatenation & 84.35 & 13.91 & 17.39 & 15.65 \\
\hline
{{Max score fusion}} & {{85.43}} & {{12.39}} & {{16.74}} & {{14.56}} \\
\hline
{{Mean score fusion}} & {{85.87}} & {{11.96}} & {{16.30}} & {{14.13}}  \\
\hline
    BP~\cite{lin2015bilinear}   & 92.28 & 7.61 & 7.83 & 7.72 \\
\hline
    CBP~\cite{gao2016compact}   & 90.87 & 8.48 & 9.78 & 9.13 \\
\hline
    FBC~\cite{FBC2020}  & 92.83 & 8.70 & 5.65 & 7.18 \\
\hline
    FBC\_FL & 93.26 & 8.91 & \textbf{4.57} & 6.74 \\
\hline
    MC\_FBC\_CE & 93.91 & 7.39 & 4.79 & 6.09 \\
\hline
    MC\_FBC & \textbf{94.57} & \textbf{6.09} &{4.78} & \textbf{5.44} \\
\hline
    \end{tabular}%
  \label{tab:fusion}%
\begin{tablenotes}
\scriptsize
\item[] FL denotes Focal Loss, and CE denotes Cross Entropy Loss.
\end{tablenotes}
\vspace{-0.3cm}
\end{table}%

The FBC method achieved better results than the traditional BP and CBP methods due to its more compact and discriminative representation based on sparse coding. More compact representations help to overcome the redundancy and burstiness issues of traditional BP. We also present the comparison results of the focal loss function in the proposed MC\_FBC scheme with the cross-entropy loss used in the original FBC. For both schemes, the focal loss function can improve the detection results in terms of classification accuracy and error rates thanks to its generalization and robustness by giving more attention to hard and misclassified examples. Overall, the proposed MC\_FBC scheme has achieved the highest accuracy and lowest error rates (except the BPCER) on the Protocol III when tested on the WFFD database.

\vspace{-0.1in}
\begin{table}[h]
\setlength{\abovecaptionskip}{0.cm}
\setlength{\belowcaptionskip}{-1.1cm}
  \centering
  \small
  \caption{{Comparisons of model size and complexity. Params: Parameter Number; FLOPs: the number of floating point operations per second.}}
    \begin{tabular}{l|ll|ll}
     \hline
    \multirow{2}[0]{*}{\textbf{Model}} &\multicolumn{2}{c|}{\textbf{VGG-16}} & \multicolumn{2}{c}{\textbf{ResNet-50}} \\
     \cline{2-5}
          & \textbf{Params} & \textbf{FLOPs} & \textbf{Params} & \textbf{FLOPs} \\
     \hline
    w/o FBC & 138.36M & 15.48G & 25.56 M    & 4.11G \\
     \hline
    w/ FBC & 15.77M & 15.78G & 27.71M & 4.52G \\
     \hline
    w/ MC\_FBC & 15.77M & 31.13G     & 27.71M & 8.63G \\
     \hline
    \end{tabular}%
  \label{tab:FLOPs}%
\vspace{-0.3cm}
\end{table}%

\begin{table*}[h]
\setlength{\abovecaptionskip}{0cm}
\setlength{\belowcaptionskip}{-1.1cm}
\renewcommand{\arraystretch}{1.2}
\setlength{\tabcolsep}{3pt}
\centering
\small
\begin{threeparttable}   %
\caption{Comparison results (\%) on the WFFD database}
    \begin{tabular}{l|cccc|cccc|cccc}
      \hline
    {\multirow{2}[0]{*}{\textbf{Method}}} & \multicolumn{4}{c|}{\textbf{Protocol I}} & \multicolumn{4}{c|}{\textbf{Protocol II}}& \multicolumn{4}{c}{\textbf{Protocol III}} \\
      \cline{2-13} 
 & \multicolumn{1}{l}{\textbf{EER}} & \multicolumn{1}{l}{\textbf{APCER}} & \multicolumn{1}{l}{\textbf{BPCER}} & \multicolumn{1}{l|}{\textbf{ACER}} & \multicolumn{1}{l}{\textbf{EER}} & \multicolumn{1}{l}{\textbf{APCER}} & \multicolumn{1}{l}{\textbf{BPCER}} & \multicolumn{1}{l|}{\textbf{ACER}} & \multicolumn{1}{l}{\textbf{EER}} & \multicolumn{1}{l}{\textbf{APCER}} & \multicolumn{1}{l}{\textbf{BPCER}} & \multicolumn{1}{l}{\textbf{ACER}}  \\
    \hline
    Multi-scale LBP~\cite{3DMAD2013spoofing}&23.50 &  27.00 &  28.50 &  27.75 & 31.15 & 36.15 &  27.69  & 31.92 &	28.91  & 31.74 &  25.65 & 28.70\\
   \hline
   Image quality~\cite{A7galba2014face} & 35.50 &  30.50  & 39.50 &  35.00& 38.85 &  39.23  &43.46  & 41.35&	41.30  & 36.96 &  43.26 &  40.11\\ 
   \hline
    Color LBP~\cite{A12boul2015face} &  31.50&  36.00&  28.50&  32.25& 	33.85&    30.77&    39.61&    35.19& 	31.52&   33.26&    34.13&    33.70\\ 
   \hline
    Haralick~\cite{agarwal2016face} &30.50& 27.50&  32.50&  30.00&	32.69&  33.08&  36.54&   34.81&	34.78&   28.04&   40.00&   34.02  \\ 
     \hline
     Recod~\cite{boulkenafet2017competition} & 17.00&  25.50&   14.50&  20.00 & 22.69&  25.77&  30.77&  28.27 &21.30&   23.91&   20.22&   22.07 \\
    \hline
     ResNet-50~\cite{tu2017ultra}&16.50 &  21.00 &  18.50  & 19.75&	17.31 &  19.23  & 21.92  & 20.58&	15.87  & 17.61& 20.43 &  19.02\\
     \hline
     VGG-16~\cite{simonyan2014very} &14.50 & 14.50  & 18.00  & 16.25&	18.08 &  13.46&   15.77 &  14.61&	14.78 &  13.48 & 17.17&  15.33\\
     \hline
    CCoLBP~\cite{peng2018ccolbp} &29.50   &26.50   &26.00  & 26.25 &28.08  & 24.62& 34.23 & 29.42 &28.04  & 26.52 &  29.13  & 27.83\\
    \hline
     Noise model~\cite{jourabloo2018face} & 31.00  & 31.00  & 48.50  & 39.75 &  41.54  & 41.54  & 41.15  & 41.35  &38.00 & 38.04  & 47.83 &  42.93\\
     \hline
     Hybrid ResNet~\cite{muhammad2019face}&8.50 &  9.00 &  13.00  &11.00 & 11.00 &  11.38  & 13.31  & 12.35 & 10.90  & 11.21  & 13.23&   12.22\\
     \hline
     Human-based &- &20.14& 11.86&  16.00 &- &32.97&  17.97&   25.47 & - &27.39& 15.31& 21.35 \\
     \hline
     {FaceBagNet~\cite{shen2019facebagnet}} & {15.28} & {19.50} & {13.00} & {16.25} & {20.33} & {20.77} & {21.92} & {21.35} & {14.60} & {17.39} & {12.39} & {14.89}  \\
     \hline
       {CDCN~\cite{yu2020searching}} & {16.50} & {17.50} & {18.00} & {17.75} & {23.48} & {26.54} & {25.38} & {25.96} & {19.22} & {19.13} & {19.57} & {19.35}\\
     \hline
      {DeepPixBiS~\cite{george2019deep} } & {6.50} & {8.50} & {4.00} & {6.25} & {8.44} & {11.54} & {5.38} & {8.46} & {7.82} & {11.52} & {4.50} & {8.04}\\
     \hline
     MC\_FBC-VGG-16 &{5.28} & {4.35 } & {6.96}  & {5.66}  &{7.34}&  {11.53} & \textbf{4.23}  & {7.88} & {5.32}& {6.09} & {4.78} &  {5.44}\\
    \hline
    MC\_FBC-ResNet-50 & \textbf{4.82}& \textbf{4.00} &\textbf{5.50} &\textbf{4.75 }& \textbf{6.55}& \textbf{7.30}& 6.54& \textbf{6.92} & \textbf{4.70}& \textbf{5.65} &  \textbf{3.91}& \textbf{4.78}\\
       \hline
\end{tabular}
\label{tab:com1}%
\vspace{-0.3cm}
\end{threeparttable}
\end{table*}

\subsection{{Model Complexity}}
{{We have compared the proposed model with the original VGG-16/ResNet-50 and FBC models in terms of model size and complexity. The comparison results within the same input face size are shown in Table \ref{tab:FLOPs}. With one more stream to obtain complementary facial color texture descriptions, the proposed MC\_FBC model has higher (almost twice) FLOPs for both VGG-16 and ResNet-50 networks than the original FBC. However, it keeps the same model size as the original FBC, which is still much smaller than that of the VGG-16 model without FBC, and only slightly higher than the ResNet-50 model. Considering the significant performance improvement as shown in Table \ref{tab:com1}, we deem MC\_FBC a good comprised solution between cost and performance.}} 

\subsection{Comparison on the proposed database}
\subsubsection{WFFD database}
Several face PAD methods were evaluated on the WFFD database, to show how they can work for super realistic 3D presentation attacks. These PAD methods have achieved promising performance in detecting 2D type or 3D mask presentation attacks. Our benchmark set includes multi-scale LBP~\cite{3DMAD2013spoofing}, image quality assessment based~\cite{A7galba2014face}, color LBP~\cite{A12boul2015face}, Haralick features~\cite{agarwal2016face}, Recod method with outstanding performance in a face spoofing detection competition~\cite{boulkenafet2017competition}, ResNet-50 based~\cite{tu2017ultra,zhang2019fine}, VGG-16 based~\cite{simonyan2014very}, Chromatic Co-Occurrence of LBP (CCoLBP)~\cite{peng2018ccolbp}, noise modeling based~\cite{jourabloo2018face}, Hybrid ResNet~\cite{muhammad2019face}, FaceBagNet~\cite{shen2019facebagnet}, CDCN~\cite{yu2020searching}, and DeepPixBiS~\cite{george2019deep}. The experimental results of all benchmark methods were obtained using the publicly available codes. In addition, we have conducted a controlled human-based detection experiment to test the ability of human vision systems in distinguishing wax figure faces from real ones. In our controlled experiment, 20 volunteers (10 men and 10 women, aged between 23 and 55) were asked to determine whether the face is real or not using our self-developed program. The classification error rates were calculated and averaged as the final result of human-based spoofing detection. 

Table \ref{tab:com1} compares the results of different face PAD schemes. For Protocol I, we can see that the existing face PAD methods for 2D or 3D mask attacks suffer from severe performance degradation with high detection error rates on WFFD, ranging from 8.5\% to 48.5\%. We attribute the poor performance to high diversity and super realistic attacks in the new database. Among them, the most learned features~\cite{boulkenafet2017competition, tu2017ultra, simonyan2014very, muhammad2019face, shen2019facebagnet, yu2020searching, george2019deep} achieved better results than hand-crafted features~\cite{3DMAD2013spoofing, A7galba2014face, A12boul2015face, agarwal2016face}. Human-based detection has achieved a lower ACER of 16\%, but with higher APCER than BPCER, suggesting that more wax figure faces were mistaken for real ones. The proposed MC\_FBC scheme achieved the best results with ACER less than 6\% for both backbone networks. Similar performance differences can be observed under Protocol II. However, most algorithms achieved higher error rates for this protocol. Such results are reasonable since recording real and wax figure faces in the same scenario with the same camera results in less difference between real and fake faces. Therefore, it is more difficult to detect spoofing attacks in this homogeneous setting. 

The overall results under Protocol III with different face PAD methods have large differences, with the error rates ranging from 3.91\% to 47.83\%. The best ACER was achieved in the proposed ResNet-50 based MC\_FBC scheme due to the highly discriminative features, which significantly outperformed other algorithms and human based detection. Based on pixel-wise binary supervision, the DeepPixBiS method~\cite{george2019deep} also achieved better results, with all error rates lower than 9\%. 
Human-based detection performs worse than machine-based for all three protocols, which implies that real vs. wax detection is nontrivial for the layperson. 
The image quality-based~\cite{A7galba2014face}, and noise modeling-based~\cite{jourabloo2018face} methods, however, performed worse on the WFFD because of the high diversity of image quality in the proposed database.

\begin{figure*}[b] 
\vspace{-0.3cm}
\setlength{\abovecaptionskip}{0pt} 
\setlength{\belowcaptionskip}{0pt} 
\centering
\subfigure[]{
\label{Fig.sub.1}
\includegraphics[height=1.80in]{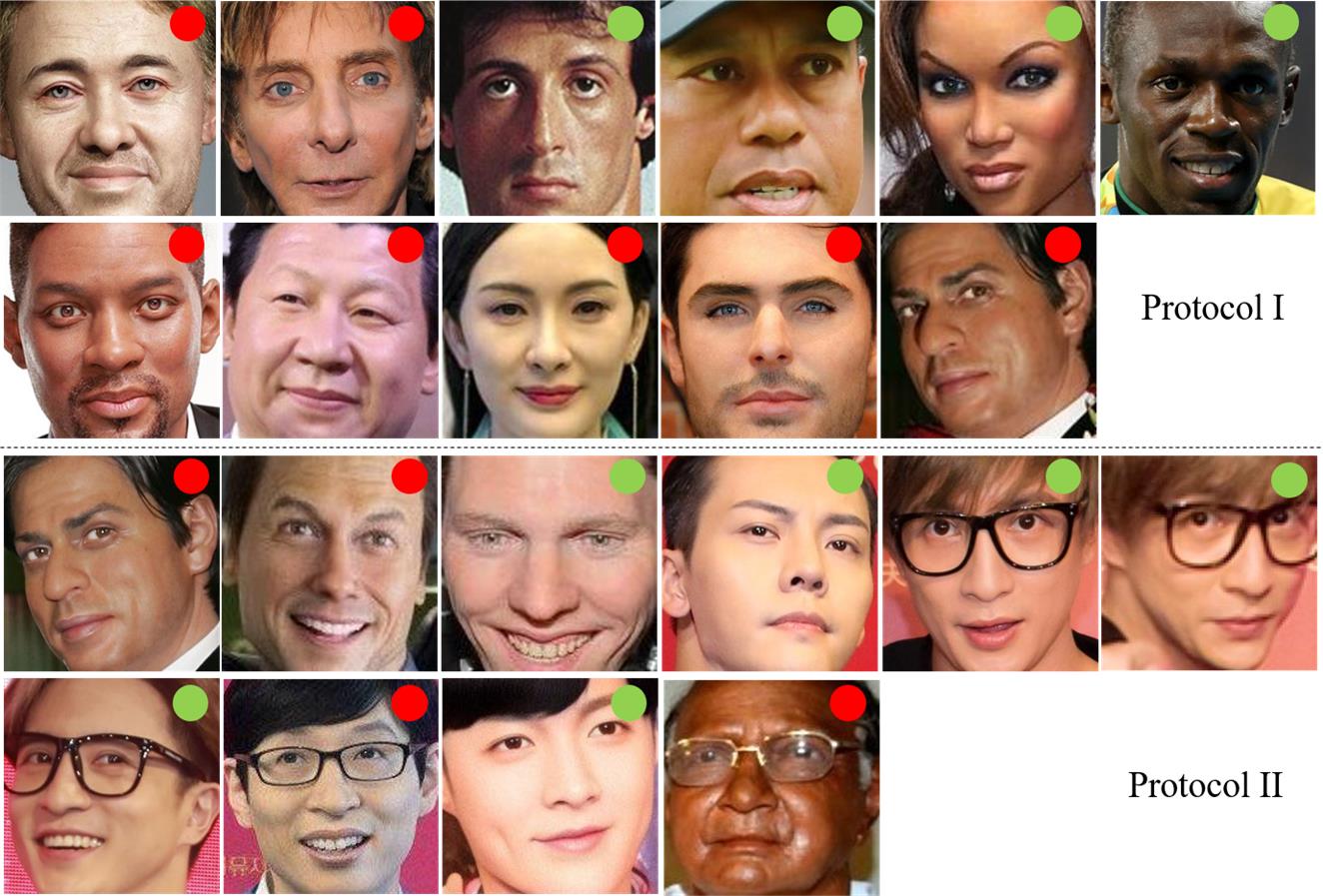}}
\hspace{0.5in}
\subfigure[]{
\label{Fig.sub.2}
\includegraphics[height=1.8in]{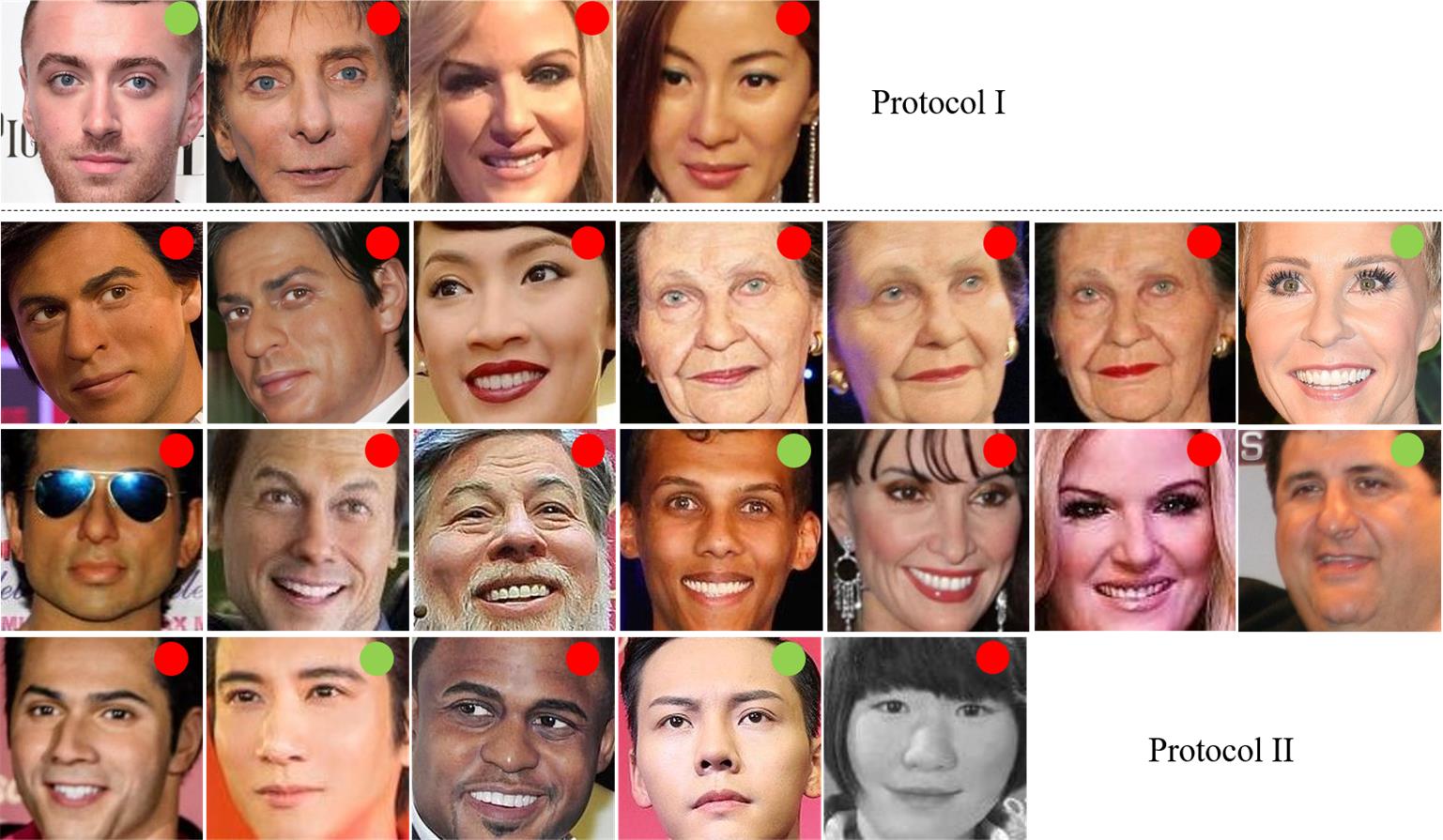}}
\caption{Failure cases with high probability. (a) in the proposed MC\_FBC method (with over 80\% of the detection results from 20 randomly chosen epochs using two backbone networks); (b) in human based anti-spoofing detection (with over 80\% of the 20 volunteers). Note that images with red dots are wax figure faces (but mistaken for real faces), while images with green dots are real faces (but mistaken for wax faces).}
\label{fig:False}
\vspace{-0.6cm}
\end{figure*} 

\subsubsection{Failure case analysis}
Based on the detection results of WFFD, we further analyze the failure cases in order to achieve a deeper understanding. In Fig. \ref{fig:False}, we have shown the failure cases with high probability in both MC\_FBC method and human-based detection results, which visually illustrate the challenges of distinguishing between fake faces and real ones even for human observers. From Fig. \ref{fig:False}(a), we note that most false detections with high probability in the proposed MC\_FBC method trend to have special face poses, which on the contrary may become the cues for human to detect. More interestingly, when compared with the proposed method, human-based detection was more likely to mistake wax figure faces for real ones for both protocols, as shown in Fig. \ref{fig:False}(b) (there are more red dots than green dots). This is in sharp contrast with that in machine-based method in Fig. \ref{fig:False}(a) (there are nearly the same number of total green and red dots).

\subsubsection{WFFD-V database}
To detect moving wax figure faces in the newly constructed WFFD-V database, we have randomly selected 10 frames from each video, and the resulting scores from the Softmax classifier were averaged to obtain the final score. Table \ref{tab:WFFDV} shows that the proposed method achieved the best performance with an accuracy of 94.74\% and ACER of 5.23\% using either VGG-16 or ResNet-50 network. The last four rows of the table demonstrate the effect of multiple color space fusion in improving the detection results. Compared with the results in Table \ref{tab:com1}, we can observe that the performance gap among different methods in detecting video-based wax figure faces is smaller than detecting photo-based ones, and most methods achieved lower error rates in this dataset. This can be attributed to the lower diversity of the WFFD-V. Overall, the learning features, especially the bilinear pooling based, patch-based~\cite{shen2019facebagnet}, and pixel-wise supervision-based~\cite{yu2020searching, george2019deep}, performed better (with accuracy over 91\% and ACER less than 9\%) than hand-crafted features. The Haralick based method~\cite{agarwal2016face} obtained the worst performance with an ACER over 20\% in distinguishing between wax figure face videos and real face ones. 

\begin{table}[t]
\setlength{\abovecaptionskip}{0.cm}
\setlength{\belowcaptionskip}{-1.1cm}
\newcommand{\tabincell}[2]{\begin{tabular}{@{}#1@{}}#2\end{tabular}}
\renewcommand\arraystretch{1.2}
  \centering
  \small
  \caption{Comparison results (\%) on the new WFFD-V database.}
  \resizebox{8.8cm}{3.8cm}{
    \begin{tabular}{lcccc}
    \hline
    \textbf{Method} & \textbf{Accuracy} & \textbf{APCER} & \textbf{BPCER} & \textbf{ACER} \\
    \hline
    Color LBP~\cite{A12boul2015face} & 84.21 & 20.69 & 10.71 & 15.70 \\
\hline
    Haralick~\cite{agarwal2016face} & 78.95 & 27.59 & 14.29 & 20.94 \\
\hline
    Recod~\cite{boulkenafet2017competition}  & 83.33 & 8.93  & 24.14 & 16.53 \\
\hline
    ResNet-50~\cite{tu2017ultra}  & 87.72 & 10.34 & 14.29 & 12.32 \\
\hline
    VGG-16~\cite{simonyan2014very} & 89.25 & 11.61 & 9.91  & 10.76 \\
\hline
    CcoLBP~\cite{peng2018ccolbp} & 80.70  & 20.69 & 17.86 & 19.27 \\
\hline
    Noise model~\cite{jourabloo2018face}  & 80.04 & 15.18 & 24.57 & 19.87 \\
\hline
    Hy-ResNet~\cite{muhammad2019face} & 89.47 & 10.34 & 10.71 & 10.53 \\
\hline
FaceBagNet~\cite{shen2019facebagnet} & 91.89 & 9.48 & 6.70 & 8.09\\
     \hline
CDCN~\cite{yu2020searching}& 92.54 & 7.33 & 7.59 & 7.46\\
     \hline
DeepPixBiS~\cite{george2019deep}& {93.42} & \textbf{6.47} & {6.70} & {6.58}\\
    \hline
    BP~\cite{lin2015bilinear}    & 91.23 & {6.90} & 10.71 & 8.81 \\
\hline
    FBC~\cite{FBC2020}   & 92.98 & {6.90} & 7.14  & 7.02 \\
\hline
    \tabincell{l}{MC\_FBC-VGG-16} & \textbf{94.74} & {6.90} & \textbf{3.57} & \textbf{5.23} \\
\hline 
    \tabincell{l}{MC\_FBC-ResNet-50} & \textbf{94.74} & {6.90} & \textbf{3.57} & \textbf{5.23} \\
\hline
    \end{tabular}}%
  \label{tab:WFFDV}%
\vspace{-0.1cm}
\end{table}%

\begin{table}[t]
\setlength{\abovecaptionskip}{0.cm}
\setlength{\belowcaptionskip}{-1.1cm}
\newcommand{\tabincell}[2]{\begin{tabular}{@{}#1@{}}#2\end{tabular}} 
\renewcommand\arraystretch{1.2}
\small
  \centering
  \caption{Comparison results (\%) on 3DMAD database}
  \begin{threeparttable}
    \begin{tabular}{p{2.1cm}ccc}
    \hline
        \textbf{Method} & \textbf{APCER} & \textbf{BPCER} &\textbf{ACER} \\
    \hline
    Haralick~\cite{agarwal2016face}&  \bm{$0.00\pm{0.0}$}     & \bm{$0.00\pm{0.0}$}      &\bm{$0.00\pm{0.0}$}  \\
     \hline
    Recod~\cite{boulkenafet2017competition} & $4.70\pm{19.4}$ &\bm{$0.00\pm{0.0}$}  & $2.35\pm{9.7}$\\
     \hline
    ResNet-50~\cite{tu2017ultra} & \bm{$0.00\pm{0.0}$}     & \bm{$0.00\pm{0.0}$}      &\bm{$0.00\pm{0.0}$} \\
    \hline
    VGG-16~\cite{simonyan2014very} & $14.21\pm{16.7}$ & $3.33\pm{8.5}$ & $9.59\pm{11.2}$ \\
     \hline
    GrPPG~\cite{li2016generalized} & -     & -     & $13.3\pm 13.3$\tnote{*} \\
\hline
    LrPPG~\cite{liu20163d} & -     & -     & $8.57\pm 13.3$ \tnote{*} \\
\hline
    \tabincell{l}{Deep dynamic\\textures~\cite{shao2018joint}}& -     & -    & 1.76\tnote{*} \\
    \hline
     BP~\cite{lin2015bilinear}& \bm{$0.00\pm{0.0}$}     & \bm{$0.00\pm{0.0}$}      &\bm{$0.00\pm{0.0}$}\\
    \hline
    FBC~\cite{FBC2020}&\bm{$0.00\pm{0.0}$}    & $1.25\pm{3.5}$&$0.61\pm 1.8$ \\
    \hline
    \tabincell{l}{MC\_FBC-\\VGG-16} & \bm{$0.00\pm{0.0}$}     & \bm{$0.00\pm{0.0}$}      &\bm{$0.00\pm{0.0}$}  \\
    \hline
    \tabincell{l}{MC\_FBC-\\ResNet-50} & \bm{$0.00\pm{0.0}$}     & \bm{$0.00\pm{0.0}$}      &\bm{$0.00\pm{0.0}$}  \\
    \hline
    \end{tabular}%
  \label{tab:3D}%
\begin{tablenotes}
\scriptsize
\item[*] Using reported results.
\end{tablenotes}
\end{threeparttable}
\vspace{-0.3cm}
\end{table}%

\begin{table}[htbp]
\setlength{\abovecaptionskip}{0.cm}
\setlength{\belowcaptionskip}{-1.1cm}
\newcommand{\tabincell}[2]{\begin{tabular}{@{}#1@{}}#2\end{tabular}} 
\renewcommand\arraystretch{1.2}
  \centering
  \small
  \caption{Comparison results (\%) on HKBU-MARs-V1+ database}
  \begin{threeparttable}
    \begin{tabular}{p{1.99cm}ccc}
    \hline
         \textbf{Method} & \textbf{APCER} & \textbf{BPCER} &\textbf{ACER}\\
    \hline
     Color LBP~\cite{A12boul2015face} & $24.00\pm{36.5}$ & $9.71\pm{9.8}$ & $16.72\pm{20.2}$ \\
    \hline
    Haralick~\cite{agarwal2016face}& \bm{$3.86\pm{7.6}$} & {$2.43\pm{3.8}$} & \bm{$3.24\pm{6.8}$} \\
    \hline
    Recod~\cite{boulkenafet2017competition} & $17.14\pm{27.2}$ & $15.71\pm{27.4}$ & $16.53\pm{28.7}$\\
    \hline
    ResNet-50~\cite{tu2017ultra} & {$17.14\pm{29.2}$}     & {$15.71\pm{26.9}$}      &{$16.43\pm{28.1}$} \\
    \hline
    VGG-16~\cite{simonyan2014very}& $16.33\pm{37.8}$ & $7.10\pm{11.2}$ & $13.66\pm{19.8}$\\
    \hline
    CcoLBP~\cite{peng2018ccolbp} & {$14.28\pm{34.1}$}     & {$12.85\pm{30.0}$}      &{$13.57\pm{31.9}$} \\
    \hline
    GrPPG~\cite{li2016generalized} & -     & -     & $16.10\pm 20.5$\tnote{*} \\
\hline
    LrPPG~\cite{liu20163d} & -     & -     & $8.67\pm 8.8$ \tnote{*} \\
\hline
    \tabincell{l}{Deep dynamic\\textures~\cite{shao2018joint}} &  -     & - & 13.44 \tnote{*}\\
    \hline
    BP~\cite{lin2015bilinear}& $9.09\pm{30.2}$  & $3.55\pm{11.8}$  &  $6.32\pm{15.6} $\\
    \hline
    FBC~\cite{FBC2020}&$ 9.09\pm 30.2 $ & ${2.64\pm{8.7}}$  & $ 5.86\pm 15.3$\\
    \hline
    \tabincell{l}{MC\_FBC-\\VGG-16} &  ${8.33\pm{28.87}}$ &$\bm{0.00\pm{0.0}}$ & ${4.17\pm{14.4}}$ \\
    \hline
   \tabincell{l}{MC\_FBC-\\ResNet-50} & ${7.27\pm{24.1}}$ &$\bm{0.00\pm{0.0}}$ & ${3.64\pm{12.1}}$ \\
    \hline
    \end{tabular}%
  \label{tab:HK}%
  \begin{tablenotes}
\scriptsize
\item[*] Using reported results. Note that the result of 'Deep dynamic textures' was conducted on a subset of HKBU-MARs-V1+.
\end{tablenotes}
\end{threeparttable}
\vspace{-0.1cm}
\end{table}%

\begin{table}[t]
\setlength{\abovecaptionskip}{0.cm}
\setlength{\belowcaptionskip}{-1.1cm}
  \centering
  \small
  \caption{Comparison results (\%) on MSU-USSA database}
  \renewcommand{\arraystretch}{1.2}
  \begin{threeparttable}
    \begin{tabular}{lccc}
\hline
    \multicolumn{1}{c}{\textbf{Method}} & \multicolumn{1}{c}{\textbf{APCER}} & \multicolumn{1}{c}{\textbf{BPCER}} & \multicolumn{1}{c}{\textbf{ACER}} \\
\hline
    Color LBP~\cite{A12boul2015face} & $3.1\pm 0.8$ & $3.0\pm 0.8$ & $3.1\pm 0.8 $\\
\hline
    Image distortion~\cite{patel2016secure} & $3.3\pm 0.7$ & $4.3\pm 2.0$ & $3.5\pm 1.0$ \\
\hline
    Haralick~\cite{agarwal2016face} & $9.1\pm0.9$ & $8.8\pm0.9$ & $8.9\pm0.9$ \\
\hline
    Recod~\cite{boulkenafet2017competition} & $3.3\pm 0.4$ & $3.4\pm 1.3$  & $3.3\pm 0.7$ \\
\hline
    ResNet-50~\cite{tu2017ultra} & $7.6\pm 0.9$ & $8.9\pm 2.6$ &  $8.3\pm 1.0$  \\
\hline
    VGG-16~\cite{simonyan2014very} & $27.7\pm 5.5$ & $27.8\pm 2.2$  & $27.8\pm 3.5$ \\
\hline
   Patch-CNN~\cite{atoum2017face} & -     & -     & $0.4\pm 0.3$\tnote{*} \\
\hline
   Depth-CNN~\cite{atoum2017face} & -     & -     & $2.2\pm 0.7$\tnote{*}\\
\hline
Two stream CNN~\cite{atoum2017face} &  -     & -    & $\bm{0.2\pm 0.2}$\tnote{*}\\
\hline 
    Deep forest~\cite{cai2019learning} & -     & -     & $1.3\pm 0.5$\tnote{*} \\
\hline
    MC\_FBC-VGG-16  & $\bm{1.0\pm 0.5}$ & $2.9\pm 1.0$ & $1.9\pm 0.5$ \\
\hline
    MC\_FBC-ResNet-50 & $1.5\pm 0.5$ & $\bm{1.6\pm 0.9}$ & $1.5\pm 0.4$\\
    \hline
    \end{tabular}%
  \label{tab:msu}%
\begin{tablenotes}
\scriptsize
\item[*] Using reported results.
\end{tablenotes}
\end{threeparttable}
\vspace{-0.3cm}
\end{table}%

\begin{table*}[t]
\setlength{\abovecaptionskip}{0.cm}
\setlength{\belowcaptionskip}{-1.1cm}
  \centering
  \caption{Comparison results (\%) on Oulu-NPU database (note that this dataset is for 2D spoofing only).}
   \resizebox{18.07cm}{3.07cm}{
  \renewcommand{\arraystretch}{1.2}
\setlength{\tabcolsep}{3pt}
\small
    \begin{tabular}{l|ccc|ccc|ccc|ccc}
\hline
    \multirow{2}[0]{*}{\textbf{Method}} & \multicolumn{3}{c|}{\textbf{Protocol I}} & \multicolumn{3}{c|}{\textbf{Protocol II}} & \multicolumn{3}{c|}{\textbf{Protocol III}} & \multicolumn{3}{c}{\textbf{Protocol IV}} \\
    \cline{2-13} 
          & \textbf{AP} & \textbf{BP} & \textbf{ACER} & \textbf{AP} & \textbf{BP} & \textbf{ACER} & \textbf{APCER} & \textbf{BPCER} & \textbf{ACER} & \textbf{APCER} & \textbf{BPCER} & \textbf{ACER} \\
\hline
Color LBP~\cite{A12boul2015face} & 5.0     & 20.8  & 12.9  & 22.5  & 6.7   & 14.6  & $14.2\pm 9.2$ & $8.6\pm 5.9$ & $11.4\pm 4.6$ & $29.2\pm 37.5$ & $23.3\pm 13.3$ & $26.3\pm 16.9$ \\
\hline
    GRADIANT~\cite{boulkenafet2017competition} & 1.3   & 12.5  & 6.9   & 3.1   & 1.9   & {2.5}   & $2.6\pm 3.9 $&${5.0\pm 5.3 }$ & $3.8\pm 2.4$ & ${5.0\pm 4.5}$ & $15.0\pm 7.1$ & $10.0\pm 5.0$ \\
\hline
    MixedFASNet~\cite{boulkenafet2017competition} & \textbf{0.0}     & 17.5  & 8.8   & 9.7   & 2.5   & 6.1   & $5.3\pm 6.7$ & $7.8\pm 5.5$ & $6.5\pm 4.6$ & $10.0\pm 7.7$ & $35.8\pm 26.7$ & $22.9\pm 15.2$ \\
\hline
    Recod~\cite{boulkenafet2017competition} & 3.3 & 13.3 & 8.3 & 15.8 & 4.2 & 10.0 & $10.1\pm 13.9 $&$ 8.9\pm 9.3$&$ 9.5\pm 6.7 $&$ 35.0\pm 37.5 $&$ 10.0\pm 4.5 $&$ 22.5\pm 18.2$\\
    \hline
    Noise model~\cite{jourabloo2018face} & 1.2   & 1.7   & 1.5   & 4.2   & 4.4   & 4.3   & $4.0\pm 1.8$ & $3.8\pm 1.2$ & ${3.6\pm 1.6}$ & $ 5.1\pm 6.3 $& $\bm{6.1\pm 5.1}$ & $\bm{5.6\pm 5.7}$ \\
\hline
    MILHP~\cite{lin2018live} & 8.3   & 0.8   & 4.6   & 5.6   & 5.3   & 5.4   & $\bm{1.5\pm 1.2}$ & $6.4\pm 6.6$ & $4.0\pm 2.9$ & $15.8\pm 12.8$ & $8.3\pm 15.7$ & $12.0\pm 6.2$ \\
\hline
    {CDCN~\cite{yu2020searching} }  & {0.4} & {1.7 } & {1.0 } & {\textbf{1.5} } & {1.4 } & {\textbf{1.5}} & {$2.4\pm1.3$ } & {\bm{$2.2\pm2.0$}} & {\bm{$2.3\pm1.4$ }} & {\bm{$4.6\pm4.6$}} & {$9.2\pm8.0$ } & {$6.9\pm2.9$ }\\
\hline
    DeepPixBiS~\cite{george2019deep} & 0.8   & \textbf{0.0}    & \textbf{0.4}   & 11.4  & \textbf{0.6}   & 6.0 & $11.7\pm 19.6$ & $10.6\pm 14.1$ & $11.1\pm 9.4$ & $36.7\pm 29.7$ & $13.3\pm 16.8$ & $25.0\pm 12.7$ \\
\hline
    TSCNN~\cite{chen2019attention} & 5.1   & 6.7   & 5.9   & 7.6   & 2.2   & 4.9   &$3.9\pm 2.8$ & $7.3\pm 1.1$ & $5.6\pm 1.6$& $11.3\pm 3.9$ & $9.7\pm 4.8$ & $9.8\pm 4.2$\\
\hline
    SAPLC~\cite{sun2020face} & \textbf{0.0}    & 0.8   & {0.4}  & {2.8}   & 2.2   & {2.5}  & $4.7\pm 4.2$ & ${3.1\pm 3.5}$ & $3.9\pm 2.1$ &  $11.9\pm 6.9$ & $6.7\pm 5.5$ & $9.3\pm 4.4$ \\
\hline
    MC\_FBC-VGG-16  & 5.7   & 8.7   & 7.2   & 5.1   & 3.3   & 4.2   & $2.9\pm 3.8$ & $9.9\pm 8.9$ & $6.5\pm 4.7$ & $6.3\pm 5.8$ & $10.5\pm 10.9$ & $8.8\pm 5.9$ \\
\hline
    MC\_FBC-ResNet-50 & 3.5   & 8.3   & 5.9   & 3.3   & 4.2   & 3.8   & $3.4\pm 3.5$ & $6.3\pm 4.2$ & $4.9\pm 3.7$ & $5.8\pm 4.5$ & $9.4\pm 9.3$ & $7.6\pm 5.4$ \\
    \hline
    \end{tabular}}%
  \label{tab:oulu}%
   \begin{tablenotes}
\scriptsize
\item[] All using reported results. `AP' denotes `APCER', and `BP' denotes `BPCER'.
\end{tablenotes} 
\vspace{-0.1cm}
\end{table*}%

\subsection{Intra-database testing on existing databases}
\subsubsection{3D mask spoofing databases}
The spoofing detection results on 3DMAD and HKBU-MARs-V1+ databases, are shown in Tables \ref{tab:3D} and \ref{tab:HK}. Besides the previous benchmark set, we include three face PAD methods proposed for 3D mask spoofing detection- namely, two heartbeat signal-based methods using global or local rPPG-spectrum features~\cite{li2016generalized, liu20163d}, and the deep dynamic texture-based method~\cite{shao2018joint}. Without publicly available codes, we have directly cited the reported results under the same protocol. It can be seen that the proposed MC\_FBC method achieved 0\% error rate on 3DMAD database, where several methods performed perfectly. Note that this is because 3DMAD is a relatively easy dataset for spoofing detection (with simple and rigid masks as shown in Fig. \ref{fig:2a}). On a more realistic HKBU-MARs-V1+ database, our method achieved 3.64\% ACER using ResNet-50 as the backbone network, slightly higher than the best result from Haralick features (with 3.24\%). We can also observe that all methods have achieved higher error rates on HKBU-MARs-V1+ (see Table \ref{tab:HK}) than on the 3DMAD database (see Table \ref{tab:3D}). Such results are reasonable because the mask spoofing samples in the HKBU-MARs-V1+ are closer to real faces and contain more realistic variations. Based on heartbeat signal analysis, the methods using GrPPG~\cite{li2016generalized} and LrPPG features~\cite{liu20163d} were affected little by the spoofing quality, showing better detection robustness on the two databases.
\subsubsection{2D face spoofing database}
Tables \ref{tab:msu} and \ref{tab:oulu} show the comparison results on the 2D face spoofing databases (MSU-USSA and Oulu-NPU). In Table \ref{tab:msu}, we also include two new methods on MSU-USSA database, including two-stream CNN in~\cite{atoum2017face} combining Patch-CNN and depth-CNN, and deep forest with multiscale LBP based method~\cite{cai2019learning}. The proposed MC\_FBC method has achieved the ACER of 1.9\% and 1.5\% using two backbone networks (VGG-16 and ResNet-50) respectively, slightly higher than deep forest based method (with ACER of 1.3\%). Using not only full face images, but also local patches extracted from the same face, the two-stream CNN method performed the best with 0.2\% ACER in distinguishing the fake from real faces. 

For the widely used Oulu-NPU database, we have added several benchmark methods, including two leading methods in face spoofing detection competition held in 2017~\cite{boulkenafet2017competition} (GRADIANT and MixedFAXNet), {{and five recent works (MILHP~\cite{lin2018live}, CDCN~\cite{yu2020searching}},  DeepPixeBiS~\cite{george2019deep}, TSCNN~\cite{chen2019attention}, and SAPLC~\cite{sun2020face})}. The proposed MC\_FBC method based on ResNet-50 model achieved ACERs of 5.9\%, 3.8\%, 4.9\%, and 7.6\% in the four protocols respectively, only slightly lower than using VGG-16 as the backbone network. The CDCN method achieved superior performance to most methods under the first three protocols with limited variations due to its powerful discrimination ability in extracting intrinsic 2D spoofing patterns~\cite{yu2020searching, yu2020multi}. 
Although the performance of our proposed MC\_FBC was not the best under Protocol I with unseen environmental conditions, it has shown good robustness (ranking third, slightly lower than the best result of noise model based method~\cite{jourabloo2018face}) under Protocol IV, which combines the previous three protocols and is the most challenging scenario. Overall, even though the proposed method is designed for 3D face anti-spoofing, it is still highly competitive when compared against state-of-the-art methods for detecting 2D face spoofing attacks.

\subsection{Inter-database testing}
To study the generalization property against unseen attacks, we have conducted inter-database evaluation on both the new WFFD and existing 3D mask spoofing databases. We first show how well existing methods can perform in detecting moving wax figure faces using the still wax figure faces as the training set. The face images in the training subset (2760) in WFFD were used for training, and all 285 videos (each with 10 frames) in WFFD-V dataset were used for testing. Table \ref{tab:cross} shows larger differences in the accuracy in the range of 52.50\% to 86.32\% and ACER of 13.72\% to 51.26\% among different detection methods. The last four rows show the effectiveness of bilinear pooling fusion on improving the performance. 

\begin{table}[h]
\vspace{-0.3cm}
\setlength{\abovecaptionskip}{0.cm}
\setlength{\belowcaptionskip}{-1.1cm}
\renewcommand\arraystretch{1.2}
  \centering
  \small
  \caption{Inter-database testing results (\%) on WFFD-V database.}
    \resizebox{8.87cm}{3.57cm}{
    \begin{tabular}{lcccc}
    \hline
    \textbf{Method} & \textbf{Accuracy} & \textbf{APCER} & \textbf{BPCER} & \textbf{ACER} \\
\hline
    Color LBP~\cite{A12boul2015face}  & 52.50  & 51.90  & 42.95 & 47.42 \\  
\hline
    Haralick~\cite{agarwal2016face} & 54.04 & 51.03 & 40.71 & 45.87 \\  
\hline
    Recod~\cite{boulkenafet2017competition} & 67.72 & 18.62 & 46.43 & 32.52 \\  
\hline
    ResNet-50~\cite{tu2017ultra} & 60.18 & 36.29 & 43.48 & 39.89 \\  
\hline
    VGG-16~\cite{simonyan2014very} & 71.23 & 38.62 & 18.57 & 28.60 \\  
\hline
    CcoLBP~\cite{peng2018ccolbp}  & 48.77 & 48.96 & 53.57 & 51.26 \\  
\hline
    Noise model~\cite{jourabloo2018face}  & 63.38 & 49.66 & 23.12 & 36.39 \\  
\hline
    Hy-ResNet~\cite{muhammad2019face} & 63.16 & 53.79 & 19.29 & 36.54 \\  
\hline
     {FaceBagNet~\cite{shen2019facebagnet} } & {67.81 } & {40.17 } & {23.93 } & {32.05} \\
     \hline
     {CDCN~\cite{yu2020searching}} & {79.91 } & {20.09 } & {20.09 } & {20.09} \\
     \hline
    {DeepPixBiS~\cite{george2019deep}} & {80.83 } & {23.10 } & {15.09 } & {19.10}\\
    \hline
    BP~\cite{lin2015bilinear}     & 77.50  & 28.79 & 15.98 & 22.39 \\  
\hline
    FBC~\cite{FBC2020}    & 78.86 & 22.59 & 19.64 & 21.11 \\  
\hline
    MC\_FBC-VGG-16 & 82.46 & 24.83 & \textbf{10.00} & 17.41 \\  
\hline
    MC\_FBC-ResNet-50  & \textbf{86.32} & \textbf{11.72} & 15.71 & \textbf{13.72} \\
    \hline
    \end{tabular}}
  \label{tab:cross}%
\vspace{-0.1cm}
\end{table}%

The proposed method demonstrates the best generalizability with the lowest ACER of 17.41\% and 13.72\% using the VGG-16 and ResNet-50 models respectively. We attribute the outstanding performance to the complementary features learned by the proposed method. Similar to the results in Table \ref{tab:WFFDV}, most learning feature based methods performed better in detecting video-based wax faces than hand-crafted features.

We have further compared the generalizability of the proposed method to unseen 3D mask spoofing attacks. For a fair comparison, we have followed the protocols in~\cite{liu2018remote, liu2020temporal}: training on 3DMAD uses random 8 subjects, training on HKBU-MARs-V1+ uses 6 subjects, and both testing uses all subjects. Besides the previous benchmark methods, the latest 3D mask PAD method based on temporal similarity of rPPG (TSrPPG)~\cite{liu2020temporal} is added to the benchmark set. Results in Table \ref{tab:intra2} have justified the robustness of the MC\_FBC method, with the second lowest ACER using ResNet-50 as the backbone under both inter-database test settings, while the VGG-16 based MC\_FBC performed slightly worse than ResNet-50 model. Most methods have achieved higher error rates using 3DMAD as the training set. The underlying reason is that this database contains less variation in the collected data than the HKBU-MARs-V1+ database. Therefore, the models optimized for this database are not able to generalize well in new acquisition conditions. Due to the good generalizability of liveness cues, the TSrPPG method~\cite{liu2020temporal} has achieved the best results using heartbeat signal in the time domain.
\begin{table*}[t]
\setlength{\abovecaptionskip}{0.cm}
\setlength{\belowcaptionskip}{-1.1cm}
\renewcommand\arraystretch{1.23}
  \centering
  \small
  \caption{Inter-database testing results (\%) on 3DMAD and HKBU-MARs-V1+ databases.}
    \begin{tabular}{l|ccc|ccc}
    \hline
    \multirow{2}[0]{*}{\textbf{Method}} & \multicolumn{3}{c|}{\textbf{3DMAD $\rightarrow$ HKBU-MARs-V1+}} & \multicolumn{3}{c}{\textbf{HKBU-MARs-V1+ $\rightarrow$ 3DMAD}}\\
\cline{2-7}          & \textbf{APCER} & \textbf{BPCER} & \textbf{ACER} & \textbf{APCER} & \textbf{BPCER} & \textbf{ACER} \\
    \hline
    Multi-scale LBP~\cite{3DMAD2013spoofing} & $45.00\pm2.9$ & $43.93\pm2.9$ & $44.46\pm2.9$  & $28.76\pm5.9$ & $28.62\pm5.8$ & $28.69\pm5.9$  \\
\hline
    Color LBP~\cite{A12boul2015face} & $40.00\pm 2.0$    & $39.29\pm 2.5$    & $39.64\pm2.2$ & $34.70\pm4.5$ & $34.26\pm4.8$ & $34.48\pm4.7$ \\
\hline
    Haralick~\cite{agarwal2016face} & $30.62\pm5.9$ & $29.64\pm5.7$  & $30.13\pm5.8$ & $21.47\pm3.8$ & $21.18\pm3.6$ & $21.32\pm3.7$ \\
\hline
    ResNet-50~\cite{tu2017ultra} &$36.00\pm8.4$ & $35.14\pm8.0$ & $35.57\pm8.2$ & $23.35\pm8.9$ & $23.06\pm9.2$ & $23.21\pm9.1$\\
    \hline
    Recod~\cite{boulkenafet2017competition} & $32.50\pm4.6$ & $31.07\pm3.9$ & $31.78\pm4.2$ & $26.35\pm1.4$ & $22.09\pm5.5$ & $24.22\pm2.7$ \\
\hline
    VGG-16~\cite{simonyan2014very} & $32.50\pm5.0$ & $31.78\pm4.5$ & $32.14\pm3.4$    & $52.71\pm3.7$  & $40.47\pm2.9$  & $48.09\pm2.4 $\\
\hline
    GrPPG~\cite{li2016generalized} & -     & -     & $46.70\pm 3.0$ & -     &-     & $31.50\pm 3.8$ \\
\hline
    LrPPG~\cite{liu20163d} & -     & -     & $39.20\pm 0.8$ & -     & -     & $40.40\pm 2.7$ \\
\hline
    TSrPPG~\cite{liu2020temporal} &-     & -     & $\bm{23.50\pm 0.5}$ & -     & -     & $\bm{16.10\pm 1.0}$ \\
\hline
    BP~\cite{lin2015bilinear}&$48.08\pm 15.2 $ &  $ \bm{19.86\pm 22.7} $ &  $  33.97\pm5.5$ &$26.54\pm9.2  $ &  $ 36.07\pm10.5 $ &  $  31.31\pm3.0$\\
    \hline
    FBC~\cite{FBC2020}&$34.75\pm9.9   $ &  $  21.19\pm18.2   $ &  $ 27.97\pm4.6$&$29.95\pm8.1$  & $25.00\pm12.0$ &  $27.50\pm3.7$  \\
    \hline
    MC\_FBC-VGG-16  & $26.25\pm0.2$  & $25.36\pm9.5$ & $25.80\pm9.4$ & $22.79\pm 3.9$ & $20.96\pm 2.0$ & $22.90\pm 4.4$ \\
    \hline
    MC\_FBC-ResNet-50  &  $\bm{25.00\pm2.0}$  & $25.78\pm3.8$ & $25.39\pm2.7$ & $\bm{18.57\pm 3.6}$ & $\bm{16.10\pm 2.6}$ & $17.34\pm 3.0$ \\
    \hline
    \end{tabular}
  \label{tab:intra2}%
\begin{tablenotes}
\scriptsize
\item[] 3D denotes `3D face reconstruction'.
\end{tablenotes}
\vspace{-0.3cm}
\end{table*}%

\subsection{Visualization Analysis}
\begin{figure}[t]
\setlength{\abovecaptionskip}{0pt} 
\setlength{\belowcaptionskip}{0pt} 
\begin{center}
\includegraphics[height=2.9in]{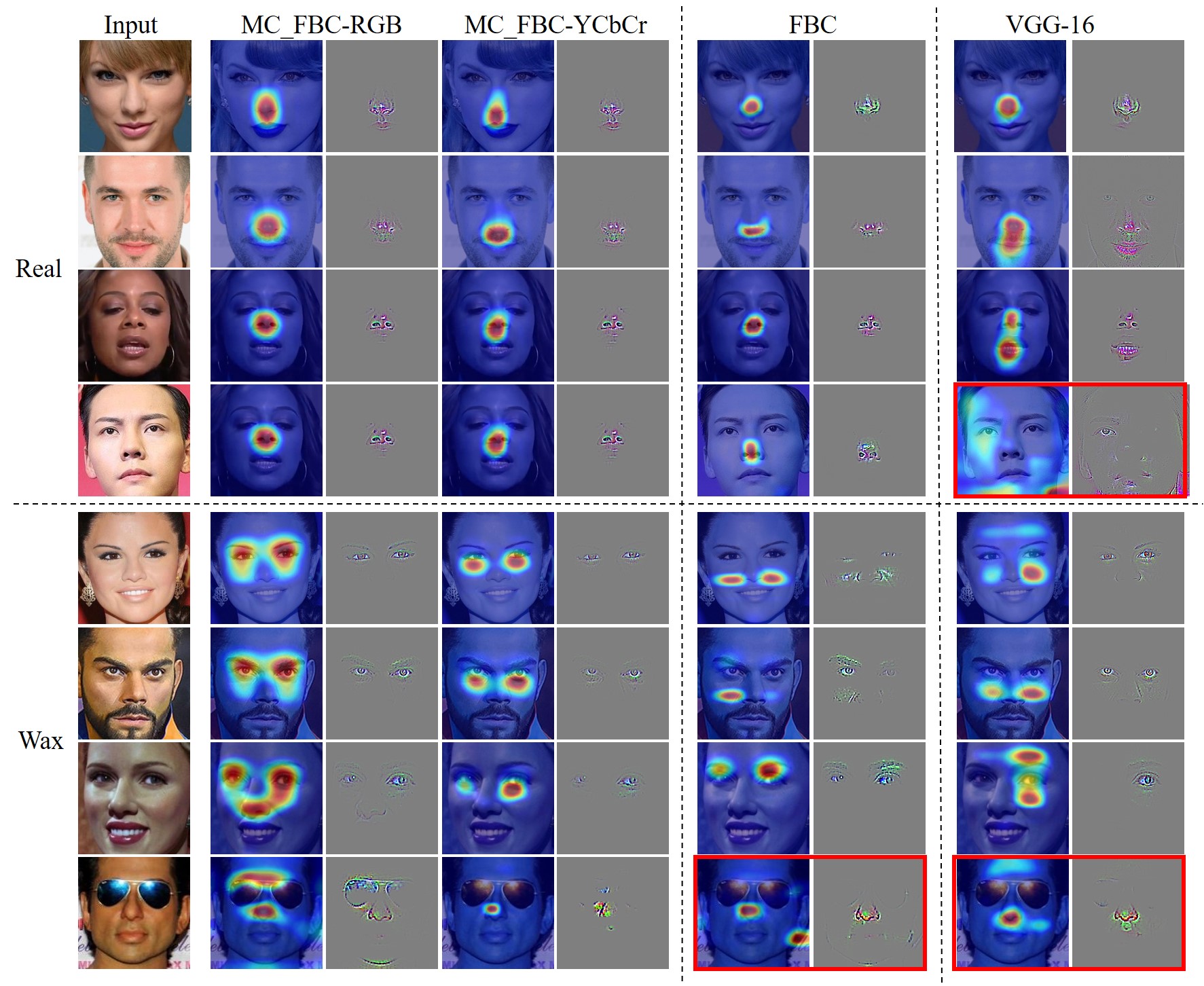}
\caption{{Grad-CAM (\cite{selvaraju2017grad}) attention visualization of the last VGG-16 feature maps corresponding to real faces (the first four rows), and wax figure faces (the last four rows) of the test data samples in WFFD. The first column represents the input faces, and the even columns show the Grad-CAM heatmaps while other odd columns show the guided Grad-CAMs. Red-colored regions of the heatmaps represent highly focused regions, whereas blue regions represent low priority ones. Note that red boxes present error detection.}}
\label{fig:gradcam}
\vspace{-0.3cm}
\end{center}
\end{figure}

To show the effectiveness of the proposed MC\_FBC on extracting highly discriminative features, we present the visualization results on WFFD samples using Grad-CAM~\cite{selvaraju2017grad}. Figure \ref{fig:gradcam} presents the CNN activation heatmaps and corresponding guided Grad-CAMs, which locate the class-discriminative regions and highlight the high-resolution details respectively. Note that Grad-CAM is not suitable for bilinear pooling~\cite{tang2020non}; thus we use Grad-CAM in the last convolution layer of VGG-16 for all models.

From the first four columns in Figure \ref{fig:gradcam}, we can first observe the complementary properties of RGB and YCbCr color spaces in the proposed MC\_FBC method. The first four rows show little difference in the attention regions of two color channels for real faces, both focusing on the nose and philtrum area. The last four rows, however, present larger attention differences in wax figure faces. We can observe from the first two rows that RGB color space generally focuses on the eye and upper cheek regions, while YCbCr color mainly focuses on the upper cheek regions. For more confusing wax samples (see the example in the seventh row with poor lighting conditions), YCbCr channel seems to be more robust due to the separation of luminance and chrominance components. However, the RGB channel performs better on wax faces with facial occlusions like eyeglasses (see the example in the last row). This can be attributed to the larger attention regions of RGB channels. Therefore, the fusion of features from complementary color channels can contribute to highly robust and discriminative representations in face anti-spoofing.

Combining with the last four columns in Figure \ref{fig:gradcam}, we can compare the class-discriminative localization of the proposed model with FBC and original VGG-16 models. It can be observed from both Grad-CAMs and guided Grad-CAMs that the high activation regions of FBC and MC\_FBC models are consistent over real faces, mainly around the nose region. Larger nose regions (with both the apex and bridge of nose) attract the attention of MC\_FBC than the FBC model (with just the apex of nose). By contrast, the VGG-16 model focuses on nose or mouth regions over real faces, while for more confusing input faces (see the example with smooth skin in the fourth row), the attention distribution scatters almost randomly, leading to erroneous detection. For the wax face samples in the last four rows, the proposed MC\_FBC model mainly focuses on eye areas and upper cheek region, the FBC model focuses on either the upper cheek or the eyebrow regions, while the focus of VGG-16 model seems to be mainly on the upper cheek regions. For the more confusing input face  (for example in the last row), both  VGG-16 and FBC lead to erroneous detection, while the MC\_FBC model is more robust thanks to the learned complementary skin-inspired features.

\section{Conclusions}
To detect realistic 3D face presentation attacks, we have proposed to generate discriminative representations in a fine-grained manner and combine complementary information in multiple color spaces by bilinear coding in this paper. The proposed MC\_FBC approach fuses complementary features from two color spaces (RGB vs. YCbCr) extracted via CNN models (VGG-16 and ResNet-50) using factorized bilinear coding. We have also constructed a new database (WFFD) with wax figure faces containing both images and videos with high diversity and large subject size as super realistic face presentation attacks. Extensive experimental results have demonstrated the superior performance of the proposed method in detecting real faces from wax figure faces with not only several existing PAD methods but also human-based spoofing detection. Our method has achieved competitive performance on other 3D mask and 2D face spoofing databases in both intra-database and inter-database testing scenarios. Both the code and databases will be made publicly available at \url{https://github.com/shanface33/Wax_Figure_Face_DB} to facilitate the improvement and evaluation of different PAD algorithms.

It should be noted that the best performance under inter-database testing achieved by the proposed scheme still has the error rate of over 10\%. How to improve the detection performance deserves further investigation. Super realistic face spoofing attacks are indeed difficult to distinguish from real ones even for humans. We envision that learning-based methods, when combined with liveness cues, are a promising direction to provide effective and generalized spoofing detection in the future. However, as AI technology keeps advancing at a fast pace, it is likely that more challenging spoofing attacks such as Deepfakes will become more powerful. 
As many people believe, the arm race between spoofing and anti-spoofing will never end. 

\section*{Acknowledgment}
Xin Li's work is partially supported by the DoJ/NIJ under grant NIJ 2018-75-CX-0032, NSF under grant OAC-1839909 and the WV Higher Education Policy Commission Grant (HEPC.dsr.18.5).

\ifCLASSOPTIONcaptionsoff
  \newpage
\fi

{\scriptsize
\bibliographystyle{IEEEtran}
\bibliography{egbib}
}





\end{document}